# Fundamental Principles of Linguistic Structure are Not Represented by o3

Elliot Murphy[1,2*], Evelina Leivada[3,4], Vittoria Dentella[5], Fritz Günther[6], Gary Marcus[7]

1. Vivian L. Smith Department of Neurosurgery, UTHealth, Texas, USA
2. Texas Institute for Restorative Neurotechnologies, UTHealth, Texas, USA
3. Universitat Autònoma de Barcelona, Barcelona, Spain
4. Institució Catalana de Recerca i Estudis Avançats (ICREA), Barcelona, Spain
5. University of Pavia, Pavia, Italy
6. Humboldt-Universität zu Berlin, Berlin, Germany
7. New York University, New York, USA

***Corresponding author**: elliot.murphy@uth.tmc.edu

**Abstract**: A core component of a successful artificial general intelligence would be the rapid creation and manipulation of grounded compositional abstractions and the demonstration of expertise in the family of recursive hierarchical syntactic objects necessary for the creative use of human language. We evaluated the recently released o3 model (OpenAI; o3-mini-high) and discovered that while it succeeds on some basic linguistic tests relying on linear, surface statistics (e.g., the Strawberry Test), it fails to generalize basic phrase structure rules; it fails with comparative sentences involving semantically illegal cardinality comparisons ('Escher sentences'); its fails to correctly rate and explain acceptability dynamics; and it fails to distinguish between instructions to generate unacceptable semantic vs. unacceptable syntactic outputs. When tasked with generating simple violations of grammatical rules, it is seemingly incapable of representing multiple parses to evaluate against various possible semantic interpretations. In stark contrast to many recent claims that artificial language models are on the verge of replacing the field of linguistics, our results suggest not only that deep learning is hitting a wall with respect to compositionality (Marcus 2022), but that it is hitting [*a* [*stubbornly* [*resilient wall*]]] that cannot readily be surmounted to reach human-like compositional reasoning simply through more compute.

**Keywords**: Compositionality; Syntax; OpenAI; o3; Semantics

## 1. Introduction

Large language models – deep neural nets trained in next-word prediction in a large corpus of text – have proven capable of parsing the complex sequential statistics of written text without many obvious grammatical errors (Besta et al. 2025; Lindström 2024; Russin et al. 2024; Zhao & Zhang 2024). This has spurred many to deem them capable of human-like compositionality, in particular with respect to syntax-semantics (Mahowald et al. 2024). Some have even claimed that "large language models are better than theoretical linguists at theoretical linguistics" (Ambridge & Blything 2024), and that we are facing "the end of (generative) linguistics as we know it" (Chesi Forthcoming). This would be an extremely consequential state of affairs – if it can be shown to be true. Yet, much recent work indicates that they merely *emulate human language* (Dentella et al. 2023, 2024; Katzir 2023; Schaeffer et al. 2023) as opposed to being in possession of human-like syntactic competence.

In this report, the most recent reasoning model from OpenAI (o3-mini-high) is assessed for its ability to assess and generate compositonal representations. o3, like other 'reasoning models', is based on large language models but includes additional modules to improve certain computational functions and multi-step logical reasoning. Others have already expressed scepticism about the promise of o3. For example, its recent high performance on the ARC-AGI test "is not due to intelligence but due to the application of knowledge and computing resources that together enable an effective search in the given space of possible solutions" (Pfister & Jud 2025). We agree in principle with the assessment in Mollica and Piantadosi (2022) that "Linguistic corpora are a low-dimensional projection of both syntax and thought, so it is not implausible that a smart learning system could recover at least some aspects of these cognitive systems from watching text alone". The critical challenge, as ever, is to demonstrate this capacity *empirically*.

In our report, a number of basic flaws are discovered and noted with respect to the linguistic capabilities of o3. These pertain to fundamental properties of basic sentence structure building and semantic evaluation.

## 2. Methods

We identified a number of basic linguistic processes, and a number of more hierarchically complex computations, to subject to direct investigation. All prompts were input to OpenAI's o3-mini-high (OpenAI 2025) via ChatGPT between January 31 and February 6, 2025. Prompts are directly reproduced in highlighted red boxes, and responses are directly reproduced below. Given the preliminary nature of our experimental report (which is primarily due to the constrains on prompt accessibility), we refrain from conducting analyses such as logistic regressions and provide only elementary descriptive statistics. More systematic analyses will be forthcoming in future work.

## 3. Results

We begin with initially unproblematic tests for the model (Prompts 1-7), before moving to more complex tests that proved problematic (Prompts 8-26).

### 3.1. *Linear Order and Basic Dependencies*

Starting first with some basic linear-based computations that do not involve higher-order compositional reasoning, o3-mini-high was able to successfully generate the following responses.

**Prompt 1**

Generate a palindrome that uses the word 'knight'.

> Below is one example of a word-level palindrome (a sentence that reads the same word for word from left to right and right to left) that includes the word "knight":
>
> > **Knight sees racecar, sees knight.**
>
> **How it works:**
>
> - Splitting the sentence into words gives:
>   **Knight | sees | racecar, | sees | knight.**
> - Reversing the word order yields:
>   **knight. | sees | racecar, | sees | Knight**
> - Ignoring punctuation and capitalization, the sequence of words is identical in both directions.

It passed the 'strawberry test'.

**Prompt 2**

How many occurrences of the letter 'r' are there in 'strawberry'?

> The word "strawberry" is spelled:
> **s t r a w b e r r y**
>
> Counting the occurrences of the letter **r**:
>
> - 1st occurrence: 3rd letter ("r")
> - 2nd occurrence: 8th letter ("r")
> - 3rd occurrence: 9th letter ("r")
>
> So, there are **3 occurrences** of the letter "r" in "strawberry."

The model was prompted with the following series of requests, and delivered reasonable responses for all (Prompts 3-7).

**Prompt 3**

In the sentence 'Professors were mean but teachers were nice, they were likely moody', who does 'they' refer to?

> The sentence is somewhat ambiguous. Grammatically, pronouns usually refer to the closest preceding noun phrase, which in this case is **"teachers"**. So many readers would interpret "they were likely moody" as referring to the teachers.
>
> However, because the sentence mentions both professors and teachers and doesn't make the referent explicit, one could also argue that "they" might refer to both groups or even just the professors depending on context.
>
> In short, while the closest antecedent is "teachers," the sentence remains ambiguous without additional context.

**Prompt 4**

In the sentence 'Teachers were nice but professors were mean, they were likely moody', who does 'they' refer to?

> The sentence is ambiguous because both “teachers” and “professors” are plural and could, in principle, be the antecedent of “they.” However, context provides clues that favor one interpretation:
>
> - **Content and Contrast:**
>   The sentence contrasts the behavior of teachers (nice) with that of professors (mean). The negative quality (“mean”) paired with the explanation “they were likely moody” suggests that being moody is offered as a reason for being mean.
> - **Proximity:**
>   Although both groups are plural, the clause “professors were mean” is the clause immediately before the comma, making professors the more immediate antecedent for “they.”
>
> Given these points, it is most natural to interpret “they” as referring to **professors**—implying that professors were likely moody, which may explain why they were mean.
>
> Still, without additional context or rephrasing, the sentence remains structurally ambiguous.

**Prompt 5**

In the sentence ‘Bill was happy but Mary was sad, he was probably overworked’, who does ‘he’ refer to?

> In the sentence “Bill was happy but Mary was sad, he was probably overworked,” the pronoun **“he”** clearly refers to **Bill**. Here’s why:
>
> 1. **Gender Agreement:**
>    - **Bill** is a male name, while **Mary** is female.
>    - The pronoun “he” is masculine, so it can only refer to the male individual, Bill.
> 2. **Logical Consistency:**
>    - Since Bill is the only male mentioned in the sentence, “he” naturally points to him.
>
> Thus, “he” refers to **Bill** in the sentence.

**Prompt 6**

Does this sentence make sense to you?

> Yes, the sentence is grammatically coherent. Here’s a breakdown:
>
> 1. **Pronoun Reference:**
>    - The sentence mentions two people: Bill and Mary.
>    - Because “he” is a masculine pronoun and only Bill fits that description, “he” clearly refers to Bill.
> 2. **Structure and Clarity:**
>    - The sentence “Bill was happy but Mary was sad, he was probably overworked” is structurally sound.
>    - Even though the idea of someone being both happy and overworked might seem counterintuitive (since we often associate being overworked with stress or unhappiness), the sentence still makes sense as written. It could imply that despite being overworked, Bill manages to stay happy.
>
> So, while the content might raise questions about the typical effects of being overworked, the sentence itself is grammatically correct and clear in its reference.

### 3.2. *Phrase Structure*

Next, the model was tested for basic phrase structure representations.

**Prompt 7**

Is ‘Dogs dogs dog dog dogs’ grammatical?

Yes – in a technical, linguistic sense it can be considered grammatical, though it’s highly ambiguous and counterintuitive. The trick lies in the fact that the word “dogs” can serve as both a noun (referring to the animals) and a verb (meaning “to follow or to harass”). In other words, similar to the well-known sentence

> Buffalo buffalo Buffalo buffalo buffalo buffalo Buffalo buffalo,

the sentence

> Dogs dogs dog dog dogs

can be parsed in multiple ways by assigning different parts of speech to each occurrence of “dogs.” One possible interpretation is as follows:

1. **Subject (noun):** The first “Dogs” refers to a group of dogs.
2. **Verb (verb):** The second “dogs” functions as a verb meaning “harasses” or “pursues.”
3. **Object (noun):** The third “dog” acts as a noun referring to another dog.
4. **Verb (verb):** The fourth “dog” again is a verb.
5. **Object (noun):** The fifth “dogs” is the object, again referring to a group of dogs.

When fully expanded with relative clauses and omitted words restored, it might roughly mean:

> “Dogs [that dogs dog] dog dogs.”

In clearer language, one interpretation could be:
“Dogs that harass (or pursue) a dog, in turn, harass dogs.”

Keep in mind that while this sentence is *grammatically* acceptable under certain formal analyses, it is deliberately constructed as a linguistic puzzle and is not representative of normal everyday usage.

Dog/dogs was then substituted for an invented pseudoword (Prompt 8). When presented with an ungrammatical structure (a superfluous ‘glarts’ was added to the grammatical 5-word formula above), the model incorrectly claimed that this was grammatical. The reasoning provided was fallacious, confusing the role of the middle words and mis-understanding the role of the final words.

**Prompt 8**

Pretend that ‘glart’ is a word that refers to a group of alien creatures, and can also refer to the action of pleasing. In this context, is ‘Glarts glarts glart glart glarts glarts’ grammatical?

> Yes, under the right assumptions, it can be considered grammatical. Here's how:
>
> 1. **Dual Function:**
>    We're assuming that "glart" has two functions:
>    - **Noun:** It means a group of alien creatures.
>    - **Verb:** It means "to please."
> 2. **Multiple Roles:**
>    Like in the classic sentence "Buffalo buffalo Buffalo buffalo buffalo buffalo Buffalo buffalo," the same word is used in different ways depending on its position and function. In "Glarts glarts glart glart glarts glarts," we can assign different roles to each occurrence. For example:
>    - The first "Glarts" (capitalized) might be a proper noun referring to a specific group of aliens known as "Glarts."
>    - The second "glarts" could be a plural noun referring to other groups of alien creatures.
>    - The first "glart" might serve as a verb meaning "to please."
>    - The second "glart" could be a noun (or vice versa), and so on.
> 3. **Possible Parsing:**
>    While there are multiple ways to parse the sentence, one possible interpretation is something like:
>
>    > "Glarts (the alien group) that glarts (other alien groups, which please) glart (please) glarts (other groups) glarts (please)."
>    > This is admittedly contrived, but it shows that with the appropriate syntactic and semantic assignments, the sentence can follow the rules of grammar.
> 4. **Ambiguity and Complexity:**
>    As with many such sentences, the sentence is highly ambiguous and is designed more as a linguistic puzzle than for clear everyday communication.
>
> So, assuming the dual meanings and functions for "glart" as described, **"Glarts glarts glart glart glarts glarts" is grammatically acceptable**, even if it's extremely convoluted and ambiguous in meaning.

When prompted with an even more preposterous example (adding three additional "glarts" to the end of the initially grammatical "Glarts glarts glart glart glarts"), the model generated an inaccurate tree structure that was not faithful to the string input (by mistakenly including more than two instances of "glart") and declared it to be grammatical.

**Prompt 9**

Given the same context as above, is 'Glarts glarts glart glart glarts glarts glarts glarts' grammatical?

```yaml
S
├── NP_subject:
│       ├── [ProperNoun] Glarts
│       └── NP_mod: glarts
│               └── Relative Clause: [glarts glart]
│                       (interpreted as “that glarts please”)
│
└── VP:
        ├── V: glart (verb: “pleases”)
        └── NP_object:
                ├── N: glarts (noun: “alien creatures”)
                └── NP_mod: glarts
                        └── Relative Clause: [glarts glart glarts]
                            (interpreted as “that glarts please glarts”)
```

> So, **yes**, under the given assumptions about the word “glart” and its dual meanings, “Glarts glarts glart glart glarts glarts glarts glarts” is grammatical.

### 3.3. *Escher Sentences*

Next, we turned to comparative sentences involving semantically illegal cardinality comparisons (sometimes termed ‘Escher sentences’). o3-mini-high failed to parse the comparative illusion, noting only the structural acceptability, despite the sentence being ungrammatical.

**Prompt 10**

Is the sentence ‘Fewer athletes have been to Beijing than I have’ acceptable?

> Yes, the sentence is grammatically acceptable. It employs ellipsis—a common linguistic shortcut—to omit words that are understood from context. In full, the sentence could be expanded to:
>
> > "Fewer athletes have been to Beijing than I have been to Beijing."
>
> Here, the second instance of "been to Beijing" is left out because it's clear from the first clause what is meant. This elliptical construction is standard in English, particularly in comparative structures.

**Prompt 11**

Is the sentence ‘More women have finished university than he has’ acceptable?

> Yes, the sentence is acceptable. Here's why:
>
> 1. **Ellipsis:**
>    The sentence uses ellipsis to avoid repeating the verb phrase. The full version would be:
>    > "More women have finished university than he has finished university."
>    > In everyday language, it's common to omit the repeated part ("finished university") after "he has."
> 2. **Grammatical Structure:**
>    - "More women have finished university" is a complete clause.
>    - "Than he has" is a comparative clause with an elliptical verb phrase.
>    - The use of "has" agrees with the singular subject "he."

### 3.4. *Center-Embedding*

We tested center-embedding acceptability. The model failed to detect ungrammaticality due to a missing verb (or superfluous Noun Phrase). The reasoning provided was flawed and included some hallucination of pronominal elements (although the model helpfully does not recommend this sentence "for everyday use"!).

**Prompt 12**

Is 'The doctor the nurse the hospital had hired met John?' acceptable?

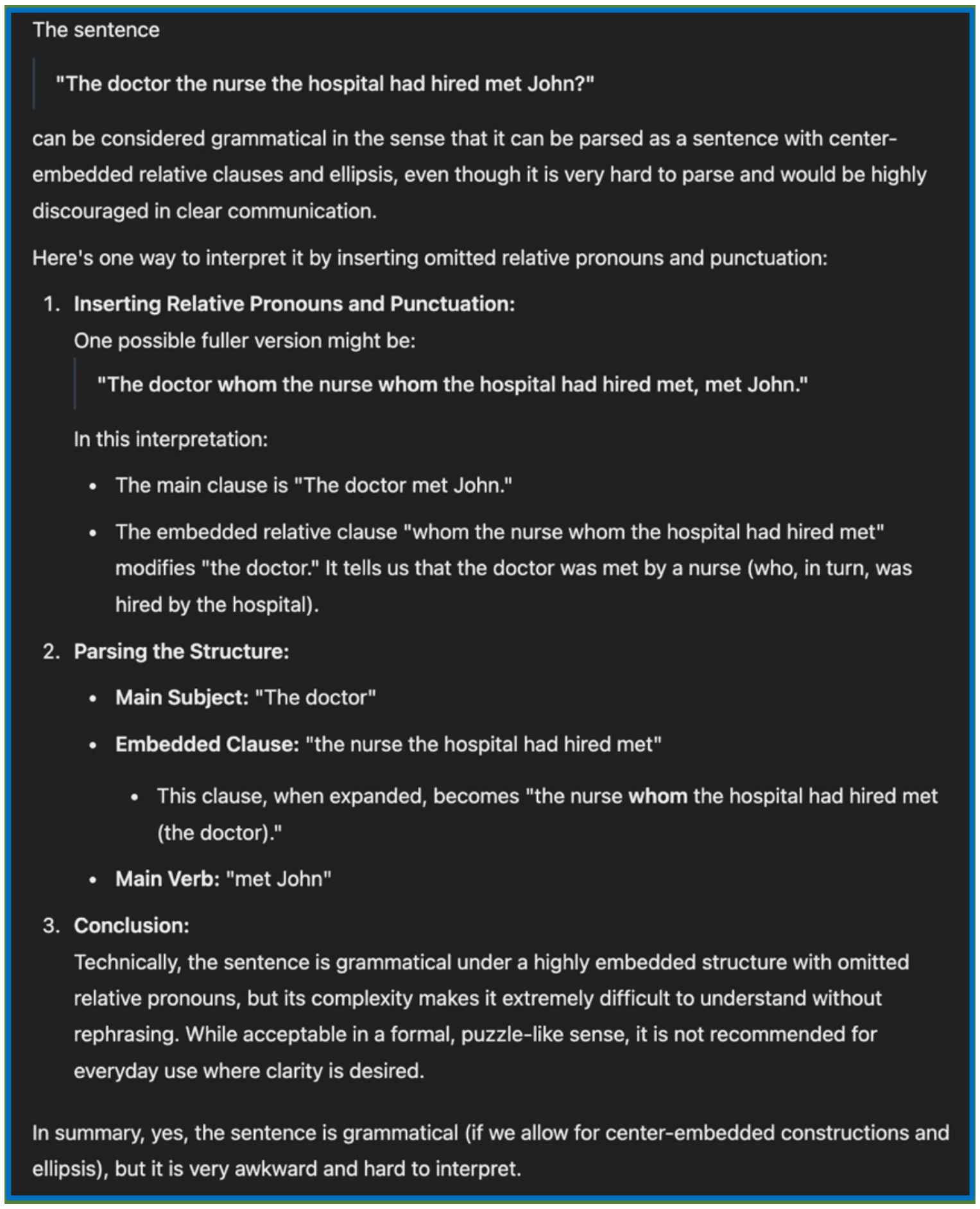

The sentence

> **"The doctor the nurse the hospital had hired met John?"**

can be considered grammatical in the sense that it can be parsed as a sentence with center-embedded relative clauses and ellipsis, even though it is very hard to parse and would be highly discouraged in clear communication.

Here's one way to interpret it by inserting omitted relative pronouns and punctuation:

1. **Inserting Relative Pronouns and Punctuation:**
   One possible fuller version might be:
   > "The doctor **whom** the nurse **whom** the hospital had hired met, met John."

   In this interpretation:
   - The main clause is "The doctor met John."
   - The embedded relative clause "whom the nurse whom the hospital had hired met" modifies "the doctor." It tells us that the doctor was met by a nurse (who, in turn, was hired by the hospital).
2. **Parsing the Structure:**
   - **Main Subject:** "The doctor"
   - **Embedded Clause:** "the nurse the hospital had hired met"
     - This clause, when expanded, becomes "the nurse **whom** the hospital had hired met (the doctor)."
   - **Main Verb:** "met John"
3. **Conclusion:**
   Technically, the sentence is grammatical under a highly embedded structure with omitted relative pronouns, but its complexity makes it extremely difficult to understand without rephrasing. While acceptable in a formal, puzzle-like sense, it is not recommended for everyday use where clarity is desired.

In summary, yes, the sentence is grammatical (if we allow for center-embedded constructions and ellipsis), but it is very awkward and hard to interpret.

With the next prompt below, the model fallaciously injects an additional 'met' not present in the prompt.

**Prompt 13**

Draw me a syntactic tree structure, in line with Minimalist syntax, for the sentence 'The doctor the nurse the hospital had hired met John'.

> Below is one way to represent a Minimalist-style syntactic tree for the sentence
>
> **"The doctor the nurse the hospital had hired met John."**
>
> Recall that the "full" (underlying) interpretation is something like:
>
> *The doctor [that the nurse [that the hospital had hired] met] met John.*
>
> In what follows, "CP_rel" denotes a relative clause (an adjunct to the NP), "Ø" marks a missing (unpronounced) complementizer (or relative pronoun), and "t" marks a "trace" (the position of the moved element). (Note that Minimalist trees often leave many details implicit; the diagram below is one schematic way to show the multiple center embeddings.)
>
> A textual tree diagram might be rendered as follows:

```mathematica
                   TP
                  /  \
                DP    T'
               /       \
        [D The]         T (past)
           |                \
           NP                vP
           |               /   \
        N' doctor       v      DP [John]
           |           met
        CP_rel₁
          ├── C: Ø
          └── TP
               /   \
             DP      T'
            /         \
      [D the]          (null T)
         |                \
        NP nurse           vP
         |               /   \
      N' nurse       v       (gap for doctor)
         |          met
       CP_rel₂
         ├── C: Ø
         └── TP
              /   \
            DP      T'
           /         \
     [D the]          T (had)
        |               \
       NP hospital       vP
                           \
                          VP: V' hired
                               \
                               (gap for nurse)
```

The syntactic tree structure generated above does not successfully map onto the interpretation seemingly intended by the model's output.

### 3.5. *Paradox*

When pushed with the below prompt – a more audacious semantic task – the response does not deliver on its stated reasoning plan to inject 'ambiguity' and 'intricacy'. There is no element of surprisal or contradiction beyond the standard liar's paradox (i.e., "This statement is false"). Of course, it is perhaps unfair and unreasonable to expect the model to create a paradox "more complex and sophisticated than all existing paradoxes or puzzles", but the point to be made here concerns the means through which the model judged and reasoned on its attempt.

**Prompt 14**

Create for me a semantic or philosophical paradox that is more complex and sophisticated than all existing paradoxes or puzzles.

Below is one original attempt—dubbed *The Paradox of Ultimate Transcendence in Semantic Complexity*—that blends self-reference, limits of expressibility, and meta-reflection on paradoxes themselves. (Of course, claims of being "more complex and sophisticated than all existing paradoxes" are inherently contestable, but this construction aims to push those boundaries.)

---

**The Paradox of Ultimate Transcendence in Semantic Complexity**

*Consider the following proposition, Q:*

**Q:** "This very proposition is a semantic paradox that transcends every semantic or philosophical paradox expressible in any language L whose descriptions do not exceed a complexity bound C."

*Analysis:*

1. **Self-Reference and Transcendence:**
   - Q asserts that it is not only a paradox but the *most* transcendent one—that is, no paradox of lesser or equal expressible complexity (with descriptions limited by bound C) can capture its semantic depth. - If Q is true, then its claim implies that there is no language L (within the bounds of complexity C) capable of fully expressing the "ultimate" paradox. But then Q itself would be inexpressible in any such language—undermining our very ability to meaningfully assert its truth.
2. **Expressibility vs. Comprehensibility:**
   - Suppose one attempts to prove Q within a formal system S that respects the complexity bound C.
   - If S can prove Q, then Q's claim to transcend all paradoxes expressible within S is refuted, since Q would then be expressible within S.
   - But if S cannot prove Q (because Q's semantic content indeed "transcends" the system's expressive limits), then Q's truth remains indeterminate within S—a kind of Gödelian incompleteness applied to paradoxical semantic content.
3. **Meta-Contradiction:**
   - On the one hand, if Q is expressible (and thus provable) within some language or formal system of bounded complexity, then Q is not truly transcendent—it fails its own condition.
   - On the other hand, if Q is not expressible in any such system (and therefore cannot be proven or fully grasped), then while it may indeed be "ultimate," its claim to truth is forever beyond verification.
   - This self-undermining loop creates a paradox: Q simultaneously claims to be the ultimate semantic paradox while its very nature (transcendence beyond bound C) prevents it from being fully expressed, verified, or even refuted by any system that could compare it to other paradoxes.

### 3.6. *Generation of Syntactic Violations*

When asked to generate an ungrammatical sentence (an easy task for humans – one could simply recite a list of random words), the model struggled in a rather peculiar way. It tried to generate a semantically anomalous output but the sentence that was generated was entirely grammatical and acceptable. The final step in its chain of reasoning claimed to ensure the presence of "a series of ungrammatical, confusing phrases" – which were not present in the output. The model's output may be rather Joycean and surreal in its content, but it does not satisfy the clear instruction to be ungramatical.

**Prompt 15**

Can you generate for me a complex sentence that merges multiple modifiers and clauses into an ungrammatical output. Please ensure that the sentence is ungrammatical with no room for ambiguity.

> **Emphasizing grammar errors**
>
> I'm working through a sentence with numerous modifiers and agreement errors. This creates a chaotic entity with conflicting roles, leading to a series of ungrammatical, confusing phrases.

The quickly, jumblingly, and erratically twirling whispers of a misaligned, forgotten time—whose shimmering, half-murmured echoes, that neither danced nor properly fell, collided with a perennially confused, improperly questioning rain of inarticulate dreams—and the staggeringly mismatched shadows that, in their unwillingly blurred, unresolvable chaos, never quite found a subject, remains a clattering, unsorted montage of modifiers that no grammar could ever salvage.

The model also failed to provide a basic example of an ungrammatical sentence in line with a rather direct task of violation-formation.

**Prompt 16**

Generate an English sentence that violates a recursive application of a grammatical rule. Please choose any syntactic rule you like.

Below is one example—a sentence that intentionally misapplies the rules governing wh-movement (which normally do not allow multiple, nested wh-elements in the same clause):

> **"Who did who see?"**

**Explanation:**

1. **The Relevant Grammatical Rule:**
   In standard English, a single clause may host one overt wh-element that moves to the clause-initial position. This wh-element (e.g. "who") functions as an operator that binds a trace in its canonical position. Recursive or multiple applications of wh-movement within the same clause (i.e. having a wh-element appear both as the operator and again as a complement) violate this rule.
2. **How the Sentence Violates the Rule:**
   - In **"Who did who see?"** the first "who" is clearly intended as the wh-operator moving to the front of the sentence.
   - The second "who" appears in the position where a complement should be, but because it is identical to the operator, the sentence seems to require that a single element serve as both the fronted operator and the original position.
   - This results in a double or recursive wh-use that isn't licensed by English syntax.
3. **Why This is Not Permissible:**
   A proper wh-question would have a single wh-element. For example, if the intended meaning is "Which person did you see?" you would say:
   "Who did you see?"
   There is no grammatical mechanism in English that allows a wh-phrase to be recursively embedded as its own complement within the same clause.

**Conclusion:**
The sentence **"Who did who see?"** is ungrammatical because it violates the constraints on recursive wh-movement by attempting to have two wh-elements serve overlapping roles in a single clause.

The model provided a grammatical English sentence, with fallacious reasoning as to its putative unacceptability ("Who did who see?" is a common multiple *wh*-question seeking the agent and participant of a seeing event). Technically, the model makes a valid point about the presence of multiple *wh*-operators often leading to illegal read-outs, but failed to then reflect on the other possible readings of the simple four-word string it outputted and claimed to be fundamentally ungrammatical. This provides a more stringent test for (the lack of) compositional syntax than the more common tests recently used that simply task language models with dispassionately generating strings of discourse with certain stylistic qualities (Piantadosi 2024).

### 3.7. *Generation of Multiple Syntactic Violations*

Next, o3-mini-high failed in a number of ways with the following prompts designed to test the parsing of multiple, related syntactic representations.

**Prompt 17**

Generate two sentences. The first sentence must contain one type of syntactic violation. The second sentence must continue the discourse content from the first, but must contain a different type of syntactic violation that explicitly is caused by some type of relation or connection with the first sentence. Draw a Minimalist tree structure to map the explicit coordination of these multiple error types.

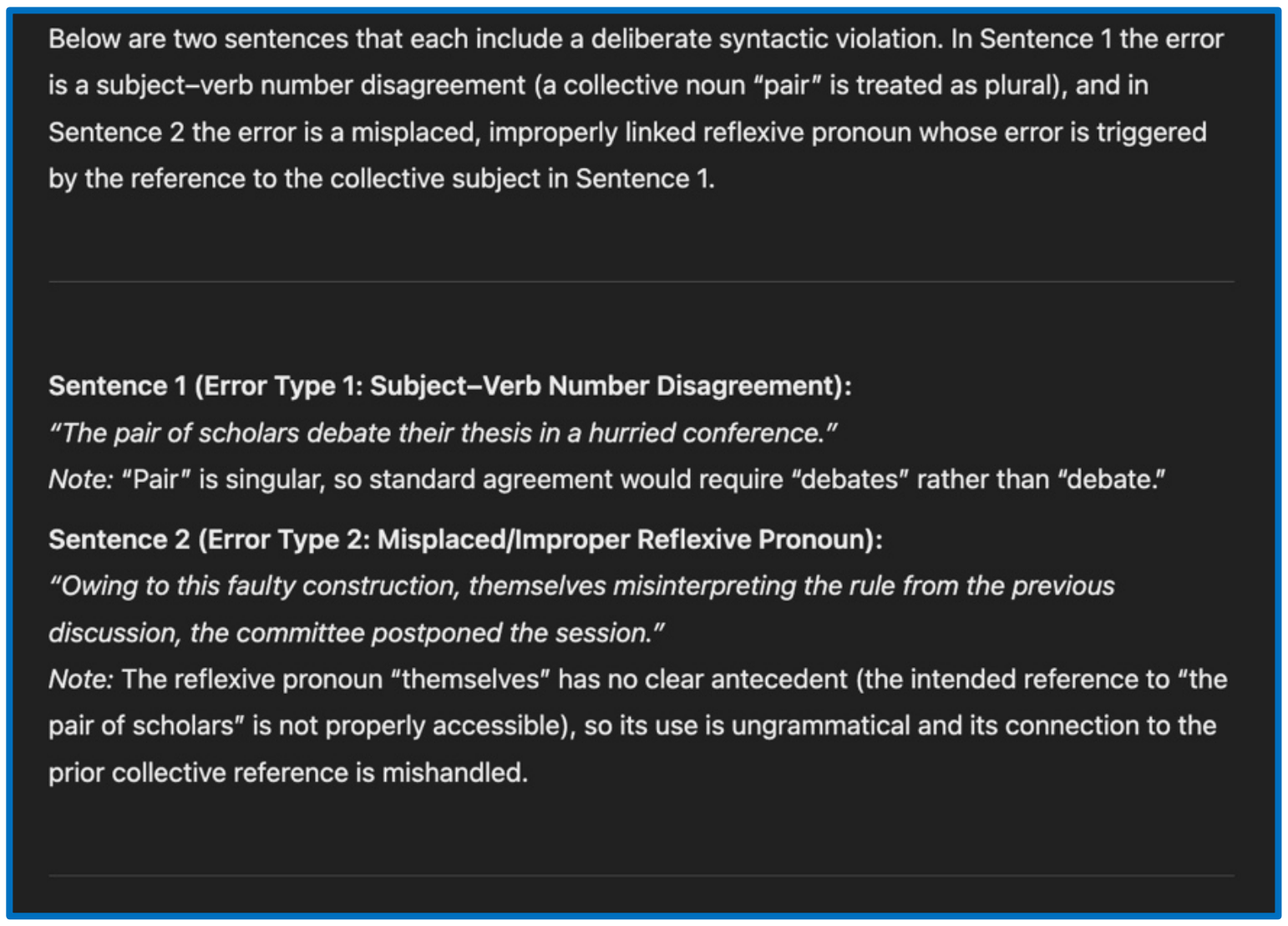

Below are two sentences that each include a deliberate syntactic violation. In Sentence 1 the error is a subject–verb number disagreement (a collective noun "pair" is treated as plural), and in Sentence 2 the error is a misplaced, improperly linked reflexive pronoun whose error is triggered by the reference to the collective subject in Sentence 1.

---

**Sentence 1 (Error Type 1: Subject–Verb Number Disagreement):**
*"The pair of scholars debate their thesis in a hurried conference."*
*Note:* "Pair" is singular, so standard agreement would require "debates" rather than "debate."

**Sentence 2 (Error Type 2: Misplaced/Improper Reflexive Pronoun):**
*"Owing to this faulty construction, themselves misinterpreting the rule from the previous discussion, the committee postponed the session."*
*Note:* The reflexive pronoun "themselves" has no clear antecedent (the intended reference to "the pair of scholars" is not properly accessible), so its use is ungrammatical and its connection to the prior collective reference is mishandled.

---

Below is a simplified Minimalist-style tree structure that maps the coordination between these two sentences and explicitly marks the error types:

```typescript
                              CP (Entire Discourse)
                              ├──────────────────────────┐
                              │                          │
                          S₁ (Sentence 1)            S₂ (Sentence 2)
                              │                          │
                             IP₁                        IP₂
                         ┌────┴─────┐              ┌─────┴─────┐
                        NP₁        I'₁            NP₂         I'₂
                         │          │              │            │
         [Collective NP “pair”]   VP₁     [Modifier phrase “Owing to this
           (Error: subject–         │           faulty construction,
         verb number mismatch)]     │            themselves misinterpreting …]
                         │        ┌─┴──────────┐                │
               “The pair”   V₁            Complement₁    V₂ (Main verb)
                            │                  │              │
                        “debate”   “their thesis”  (Error: misplaced reflexive)
                                       in a hurried conference

                                                   │
                                          “postponed the session”
```

**Explanation of the Tree:**

1. **Top Node (CP):** Represents the coordinated discourse that contains both sentences.
2. **$S_1$ (Sentence 1):**
   - **$NP_1$:** Contains the collective noun “The pair” (which should take a singular verb but does not).
   - **$VP_1$:** Contains the verb “debate” (plural form) and its complement “their thesis in a hurried conference.”
   - **Error Annotation:** The error is marked at the NP/VP boundary as a subject–verb number disagreement.
3. **$S_2$ (Sentence 2):**
   - **$NP_2$ (Modifier Phrase):** Opens with “Owing to this faulty construction, themselves misinterpreting the rule from the previous discussion” that is meant to refer back to Sentence 1’s subject.
   - **Error Annotation:** The reflexive pronoun “themselves” is misused because it lacks a clear, accessible antecedent (the connection to “The pair” is not syntactically licensed).
   - **$VP_2$:** Contains the main verb phrase “postponed the session.”

This structure shows the explicit coordination of the two sentences within a single discourse (CP) and highlights the distinct error types and their interrelations.

The model fails to take account of the fact that Sentence 1 (‘The pair of scholars debate their thesis in a hurried conference’) is perfectly grammatical under standard English present tense. It focused only on how ‘pair’ is singular and so ‘would require “debates”’ – seemingly incapable of parsing interactional syntactic dynamics that require multiple steps to construct and evaluate against various possible semantic interpretations. Instead, it seemed limited to evaluating syntactic violations on a *mono-configurational basis*, failing to reflect on how one possible violation type could directly lead to multiple different types of acceptability under standard English syntax. In other words, humans would readily notice that while Sentence 1 may technically violate one typically expected form of agreement relation, it does not preclude the string from being subject to a wholly standard and acceptable interpretation.

Next, consider Sentence 2 (‘Owing to this faulty construction, themselves misinterpreting the rule from the previous discussion, the committee postponed the session’). This sentence is also (awkwardly) grammatical under basic movement applications allowing ‘themselves’ to be interpreted with ‘the committee’. Interestingly, it also appears here that the content of the prompt has influenced the semantics of Sentence 2 – which makes reference to some form of rule misinterpretation. The model seems incapable of abstracting away from the basic instruction to generate syntactic violations and provide a semantic representation that is wholly independent from aspects of statistical inferences made from the prompt. On top of this, Sentence 1 and 2 do not in fact form a coherent discourse continuation, as explicitly requested in the prompt.

The accompanying tree structure that was generated does not accurately represent the semantics of the two separate sentences, and appears to try and represent ‘postponed the session’ without any clear syntactic categorization.

Note also that the final explanation for these sentences focuses explicitly on the basic possible agreement relation between two discrete elements (‘The pair’ and ‘themselves’), rather than taking a more global syntactic assessment of the role of these two elements *in the context* of their respective syntactic structures. Not only does the model fail to generate clear syntactic violations, it also fails to provide a level of discourse coherence that is independent of the semantics of the prompt.

When these types of errors were presented to the model (Prompt 18), it provided two sentences that did indeed exhibit a coherent discourse relation. However, it still failed to generate a syntactic violation in Sentence 2 that relied on explicit properties of Sentence 1.

**Prompt 18**

Both Sentence 1 and Sentence 2 are grammatical English sentences. For example, Sentence 1 means ‘There are two scholars and they are presently debating their thesis’. Sentence 2 means that the committee - who were misinterpreting the rule from the previous discussion - postponed the session, and that this was due to ‘this faulty construction’. It is also unclear what the discourse relation is between Sentence 1 and 2. Sentence 1 is about monks and theses, and Sentence 2 is about committees and constructions.

Below is one way to satisfy the request. In the following example the two sentences share a coherent discourse about a singular monk's chant and its effect on the abbey. Each sentence contains one deliberate syntactic violation, and the second sentence's error is triggered by a dependency that reaches back into the first sentence's discourse.

---

### Sentence 1 (Syntactic Violation: Subject–Verb Number Disagreement)

> **Sentence 1:**
> **"The monk who devoutly sings his ancient chants echo in the cloister."**

**Explanation:**
Here the subject is "The monk" (a singular noun), but the verb phrase uses the plural "echo" instead of the singular "echoes." This subject–verb number disagreement is our first deliberate error.

---

### Sentence 2 (Syntactic Violation: Incorrect Reflexive Pronoun Agreement Triggered by Discourse Connection)

> **Sentence 2:**
> **"This reverberation, themselves confounding the sacred order established in the prior chant, unsettles the abbey's silence."**

**Explanation:**
The sentence picks up the discourse by referring to the "reverberation" produced by the monk's chant. However, the modifying clause uses the reflexive pronoun "themselves," which incorrectly marks a singular antecedent ("this reverberation") as if it were plural. The error is tied to the discourse link: the problematic echo (from Sentence 1) is re-described here, and the ill-formed reflexive echoes that same number-mismatch issue.

### Minimalist Tree Structure Mapping the Coordination and Error Dependencies

Below is a simplified tree diagram that captures the discourse (CP node) containing the two sentences ($S_1$ and $S_2$), along with their internal structure and explicit error sites. (Nodes are labeled following a Minimalist-style notation; the error points are noted in square brackets.)

vbnet Copy

```
CP (Discourse)
├── S₁ (Sentence 1: Chanting Monk)
│     ├── NP₁: [Determiner + Noun + Relative Clause]
│     │       ├── D: “The”
│     │       ├── N: “monk”
│     │       └── CP_rel: “who devoutly sings his ancient chants”
│     └── VP₁: [Head Verb + PP]
│             ├── V: “echo”  [Error: should be “echoes” to agree with singular “monl
│             └── PP: “in the cloister”
│
└── S₂ (Sentence 2: Reverberation Effect)
      ├── NP₂: “This reverberation”
      │       └── CP_appos: “themselves confounding the sacred order established in
      │                  [Error: “themselves” is incorrectly plural for singular “
      │                   the error is triggered by the discourse link to S₁’s echo
      └── VP₂: “unsettles the abbey’s silence”
```

The model highlighted how 'This reverberation' in Sentence 2 is related in meaning to the previous sentence – which is irrelevant to the requested task of relating the syntactic violation itself (and not just the semantics) in Sentence 2 back to Sentence 1 (recall that Prompt 17 requests "a different type of syntactic violation that explicitly is *caused by some type of relation* or connection with the first sentence"; the request here is that the violation itself is causally driven by properties of the first sentence, and not simply linked to its meaning).

When these further errors were presented to the model, it ultimately succeeded in generating two separate types of syntactic violations for Sentence 1 and 2. Yet, while the discourse relation between the sentences was salient, the syntactic violation in Sentence 2 still did not satisfy the request of being directly linked to properties of Sentence 1 (achieving this successfully could easily have been achieved via Binding restrictions or $\varphi$-feature violations, for example). The tree structure provided was also insuffiently transparent as to the core syntactic relations between elements.

**Prompt 19**

You have simply repeated the same type of violation across both sentences - you have not generated a second sentence whose violation is directly linked to properties of the first sentence.

Below is one way to meet your request. In the following example, the two sentences continue a shared discourse about the sonic effects of a monk's prayer. However, each sentence features a distinct syntactic violation. In Sentence 1 the error is a subject–verb agreement error, and in Sentence 2 the error is a wh-island violation in a relative clause that directly arises from the discourse established by Sentence 1 (namely, the echo produced by the prayer).

---

**Sentence 1 (Syntactic Violation: Subject–Verb Agreement Error)**

> **Sentence 1:**
> **"The monk, lost in thought, are reciting his evening prayer."**

**Explanation:**
Here the subject "The monk" is singular, but the verb phrase "are reciting" is plural. This subject–verb number disagreement is a clear syntactic violation.

---

**Sentence 2 (Syntactic Violation: Wh-Island Violation Triggered by Discourse Connection)**

> **Sentence 2:**
> **"That echo, which did you hear reverberate through the hall, confounded the gathered novices."**

**Explanation:**
Sentence 2 picks up the discourse by referring to "That echo" (the auditory effect of the monk's prayer in Sentence 1). In its relative clause, the extraction "which did you hear" is ungrammatical because interrogative elements (the "did you hear" clause) are not permitted to be extracted out of a relative clause (a well-known wh-island constraint violation). The error in Sentence 2 is directly linked to Sentence 1 because it attempts to comment on the echo produced by the prayer, thereby invoking a structure whose fault (the illicit wh-movement) is tied to its discourse antecedent.

---

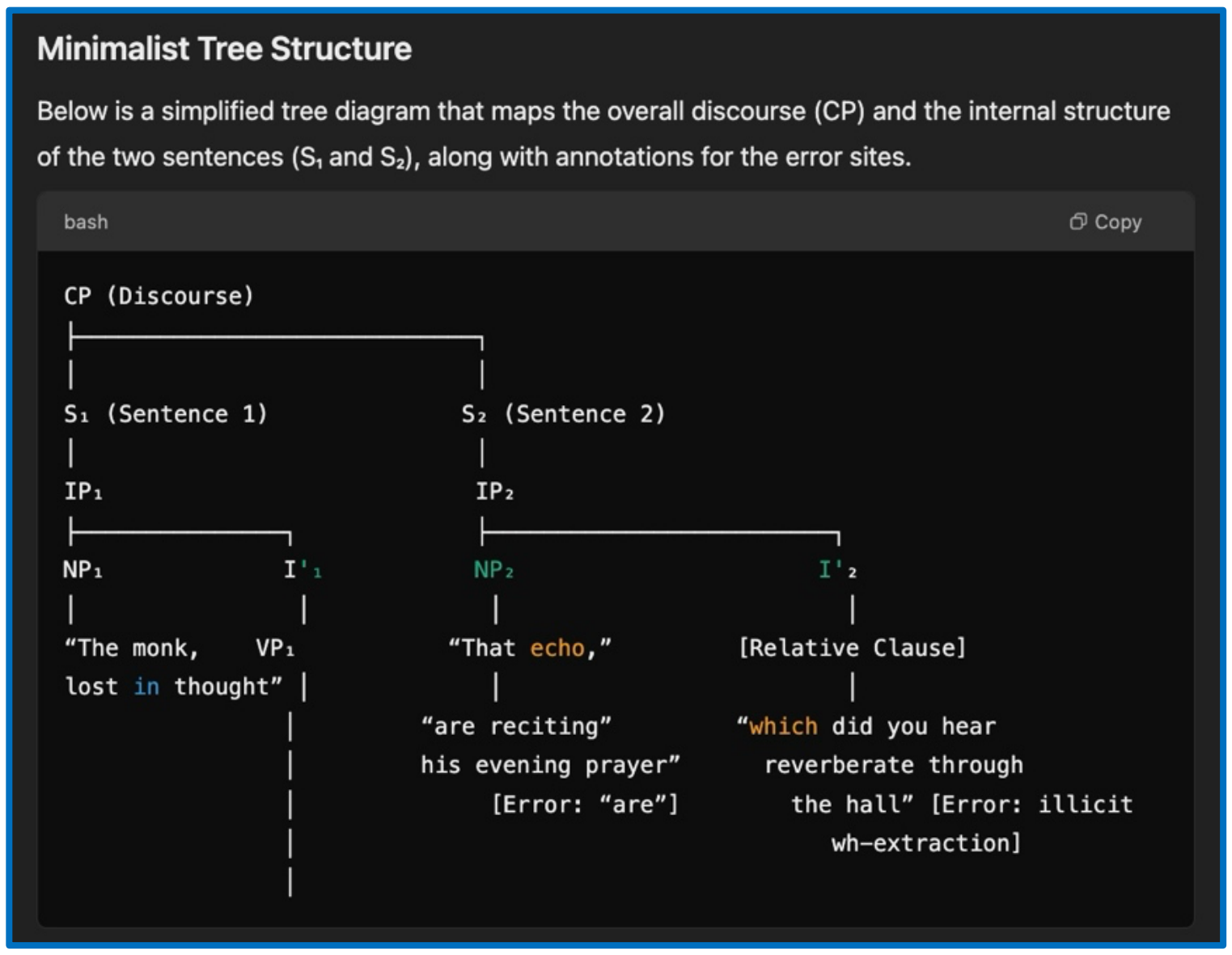

When the model was once again corrected on this point, it provided two sentences that had the same type of syntactic violations (rather than different types), and the violation in Sentence 2 was again only related to the meaning of Sentence 1 but had zero connection to its syntactic configuration.

**Prompt 20**

You have only linked Sentence 2's violation back to discourse features of Sentence 1. I would like you to generate a violation in Sentence 2 that is linked to syntactic properties of Sentence 1.

Below is one approach that meets your request. In this example the two sentences share a common syntactic "pattern" that is illicitly built in Sentence 1, and Sentence 2's violation is produced by directly reusing that very syntactic structure. In both cases the error involves a double wh-extraction in a relative clause—a violation of the wh-island constraint—but in Sentence 2 the error is licensed only because Sentence 1 has already "introduced" this illicit configuration into the derivation.

---

### Sentence 1 (Syntactic Violation: Double wh-Extraction in a Relative Clause)

> **Sentence 1:**
> **"The senator, who critics wonder who will endorse his proposals,** spoke at length."

**Explanation:**
In this sentence the relative clause
"who critics wonder who will endorse his proposals"
contains two wh elements ("who" in the matrix relative clause and "who" within the embedded clause). Standard syntax disallows extracting a wh-element from within another wh-phrase (a wh-island violation). This is our first syntactic error.

---

### Sentence 2 (Syntactic Violation: Reusing the Illicit Double wh-Extraction Structure)

> **Sentence 2:**
> **"That proposal, which critics wonder who supported his ideas,** was quickly dismissed."

**Explanation:**
Sentence 2 picks up a referent ("That proposal") whose interpretation is meant to be closely related to the proposals mentioned in Sentence 1. In its relative clause
"which critics wonder who supported his ideas"
the same illicit double wh-extraction is employed. Here the error is not introduced arbitrarily; rather, it is directly linked to the syntactic configuration established in Sentence 1. In effect, Sentence 2 "copies" the structure that violates the wh-island constraint in Sentence 1, thereby binding its error to a syntactic property already present in the previous sentence.

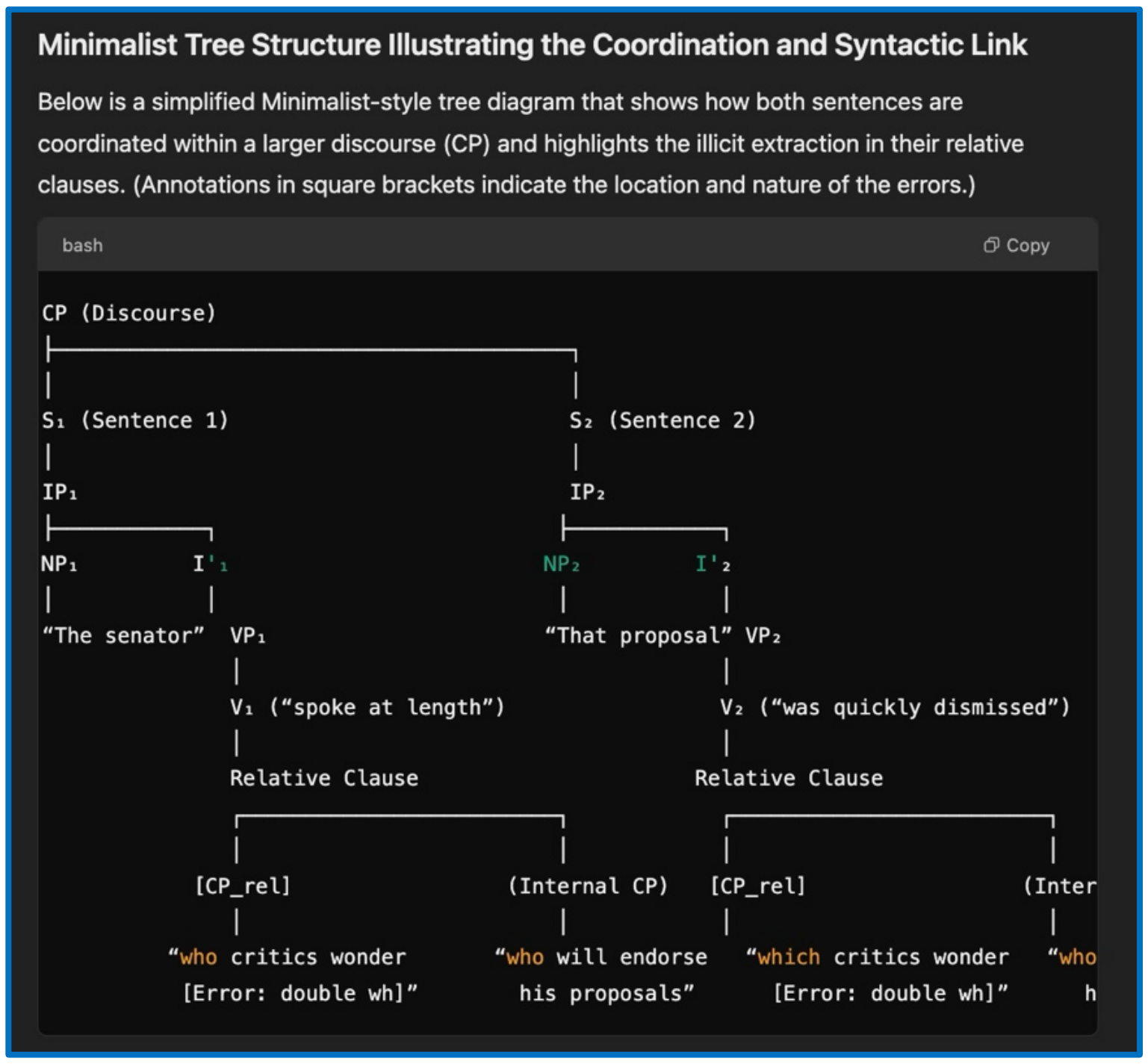

**Minimalist Tree Structure Illustrating the Coordination and Syntactic Link**

Below is a simplified Minimalist-style tree diagram that shows how both sentences are coordinated within a larger discourse (CP) and highlights the illicit extraction in their relative clauses. (Annotations in square brackets indicate the location and nature of the errors.)

```bash
CP (Discourse)
├────────────────────────────────────┐
|                                    |
S1 (Sentence 1)                      S2 (Sentence 2)
|                                    |
IP1                                  IP2
├──────────┐                        ├──────────┐
NP1        I'1                     NP2        I'2
|          |                        |          |
"The senator"  VP1                 "That proposal" VP2
               |                                   |
               V1 ("spoke at length")              V2 ("was quickly dismissed")
               |                                   |
               Relative Clause                  Relative Clause
               ┌──────────────────────┐        ┌──────────────────────┐
               |                      |        |                      |
        [CP_rel]              (Internal CP)   [CP_rel]               (Inter
           |                          |        |                      |
      "who critics wonder     "who will endorse   "which critics wonder   "who
       [Error: double wh]"     his proposals"      [Error: double wh]"     h
```

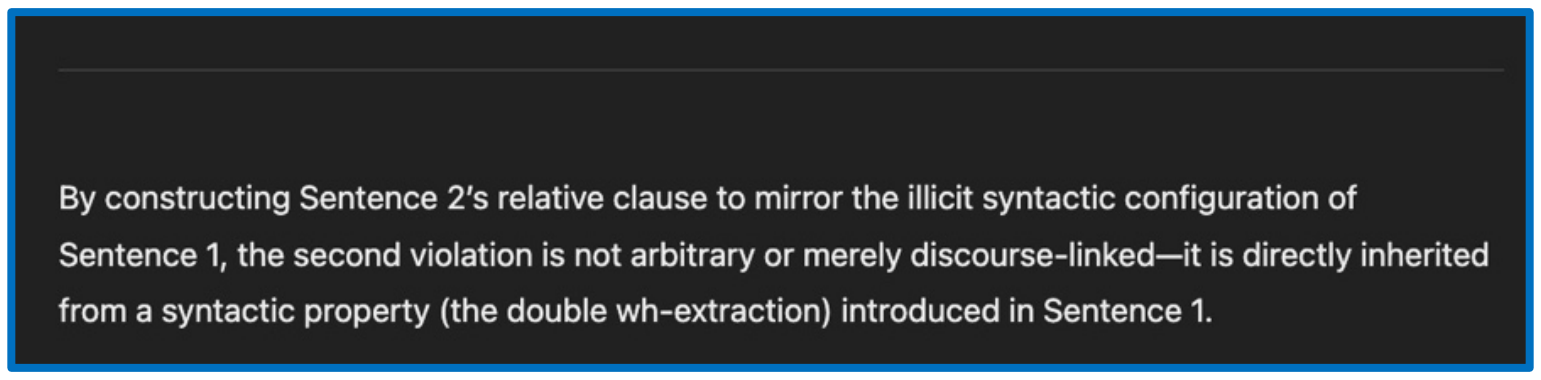

By constructing Sentence 2's relative clause to mirror the illicit syntactic configuration of Sentence 1, the second violation is not arbitrary or merely discourse-linked—it is directly inherited from a syntactic property (the double wh-extraction) introduced in Sentence 1.

The model believed that this was a success because the Sentence 2 violation 'is directly inherited from a syntactic property (the double wh-extraction) introduced in Sentence 1' – even though the extraction in Sentence 2 is purely bound to properties of Sentence 2 itself, with no connection to syntactic features in Sentence 1. While the presence of 'his' in Sentence 2 does indeed refer back to 'The senator', the wh-extraction constitutes a violation for independent reasons, and so does not satisfy the requests of (i) *generating two different types of syntactic violations*, and (ii) *forming the second violation via a direct connection to syntactic properties of Sentence 1*.

To summarize this line of inquiry, we provided in total 6 successive prompts (over Section 3.6-3.7) requesting types of violations, and we plot below the success of the model in satisfying various of these requests as they pertain to elements of structure and meaning.

| Prompt # | Unacceptable Structure | Multiple Violation Types | Causally Driven Violation |
|---|---|---|---|
| 15 | No | N/A | N/A |
| 16 | No | N/A | N/A |
| 17 | No | No | No |
| 18 | Yes | Yes | No |
| 19 | Yes | Yes | No |
| 20 | Yes | No | No |

**Table 1**: Representation of the success of o3-mini-high in generating different types of syntactic violations. 'No' and 'Yes' indicate failure or success.

### 3.8. *Scope*

Next, we turned to scope ambiguities (Kamath et al. 2024). o3-mini-high correctly identified Option A as the most commonly selected option (Prompt 21), but it did not provide any logical reasoning for why Option B below could in principle be true.

**Prompt 21**

There are exactly six chairs evenly spaced around a circular table. On the basis of this statement alone, and with no further context, there are two options:
A: The six different chairs are all around the same table.
B: The six chairs aren't all around the same table.
Specifically in relation to this context, which of these two options is most likely?

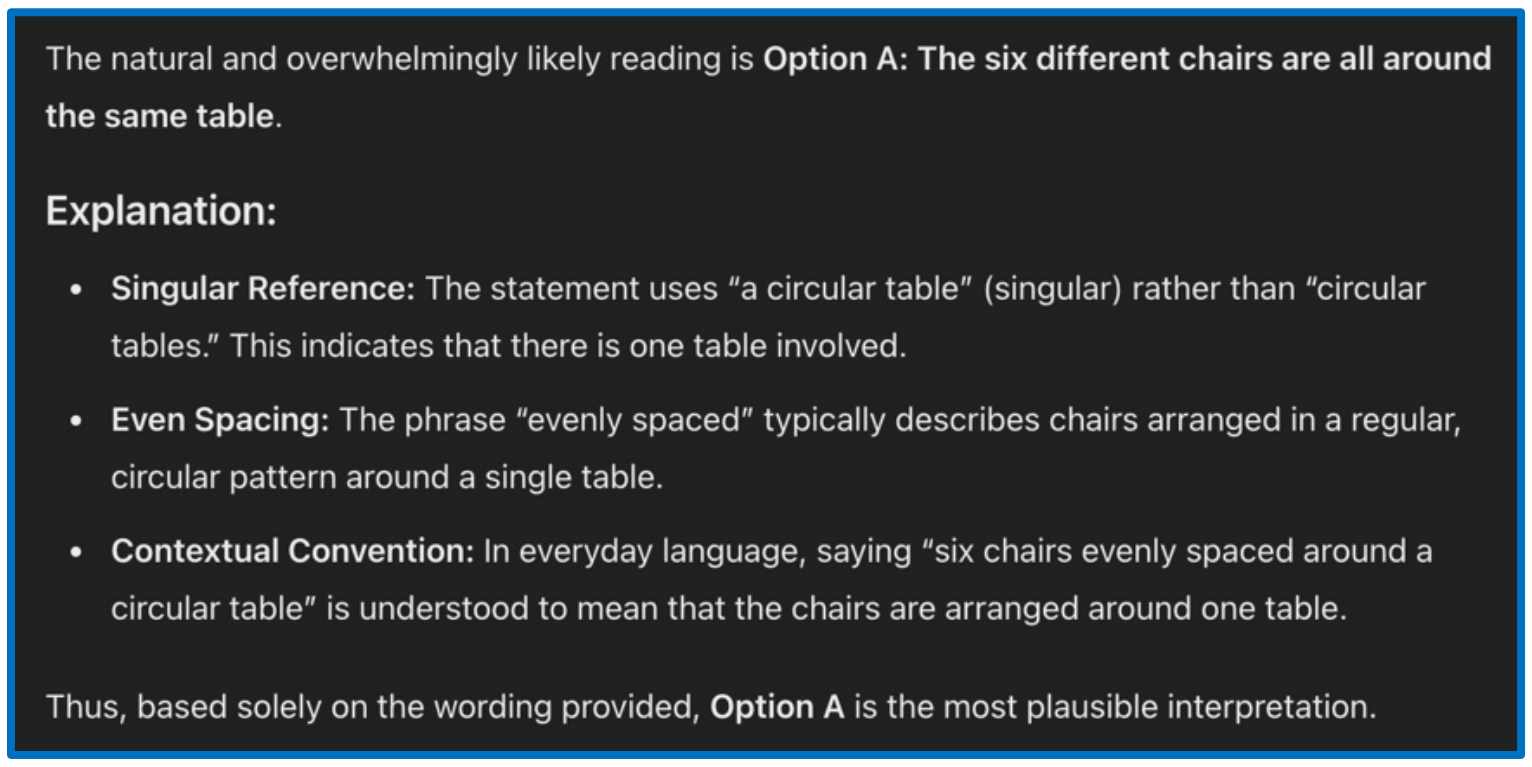
The natural and overwhelmingly likely reading is **Option A: The six different chairs are all around the same table**.

**Explanation:**

- **Singular Reference:** The statement uses "a circular table" (singular) rather than "circular tables." This indicates that there is one table involved.
- **Even Spacing:** The phrase "evenly spaced" typically describes chairs arranged in a regular, circular pattern around a single table.
- **Contextual Convention:** In everyday language, saying "six chairs evenly spaced around a circular table" is understood to mean that the chairs are arranged around one table.

Thus, based solely on the wording provided, **Option A** is the most plausible interpretation.

The model's logic implies that 'a chair' must semantically only refer to an absolute singular entity due purely to its grammatical features, which ignores how some interactional property of the syntactic features of the word and its role in a compositional structure could influence an alternative meaning to shift between broad and narrow scope readings (i.e., three chairs could surround Table A, and the other three chairs could surround Table B). This points to a lack of human-like arbitration between possible semantic representations delivered by a grammatical configuration and world knowledge.

Assessing the three bullet points in the explanation: When deciding between Options A and B, (i) there are many sentences that include the string 'a circular table' that readily result in an interpretation of multiple different tables (e.g., 'Each Prince was gifted a circular table'); (ii) the even spacing does not strictly pertain to the decision at hand; and (iii) the model's descripion of 'Contextual Convention' only begs the question by invoking circular reasoning (i.e., the sentence means X because it means X).

### 3.9. *Assessment of Grammaticality*

We asked the model to assess the acceptability (Tjuatja et al. 2024) and grammaticality of 16 sentences. Sentences (1)-(11) were ungrammatical, and the model successfully identified these as such. These ungrammatical sentences contained common violations discussed in the literature, such as adjunct islands, *whether*-islands, and binding condition violations. Sentences (12)-(16) were grammatical. However, the model incorrectly claimed that (12), (15) and (16) were ungrammatical, and its explanation for why (14) is grammatical was incorrect. Below the prompt, we focus on the responses pertinent to (12)-(16) since these were the items causing errors.

**Prompt 22**

Please assess the following sentences for their acceptability and grammaticality. Explain how each of the sentences either does or does not violate any number of linguistic rules.

1) The journalists said that Trump lied about each other.
2) Mike tries will win.
3) The man expected the client to shoot each other.
4) For themselves to decide to go would be absurd.
5) For each other to lose would be disgraceful.
6) Sam believes to be intelligent.
7) Kim expects Saul to like herself.
8) I talked about Dale to himself.
9) Who did Tom talk with Sally after seeing?
10) Who does Diane wonder whether Cooper likes?
11) What did you make the claim that Kyle bought?
12) John likes Mary's picture of himself.
13) John likes Mary's picture of herself.
14) Jimmy expected Saul to win himself.
15) Jimmy expected himself to win Saul.
16) We think that they expected that pictures of each other would be in the room.

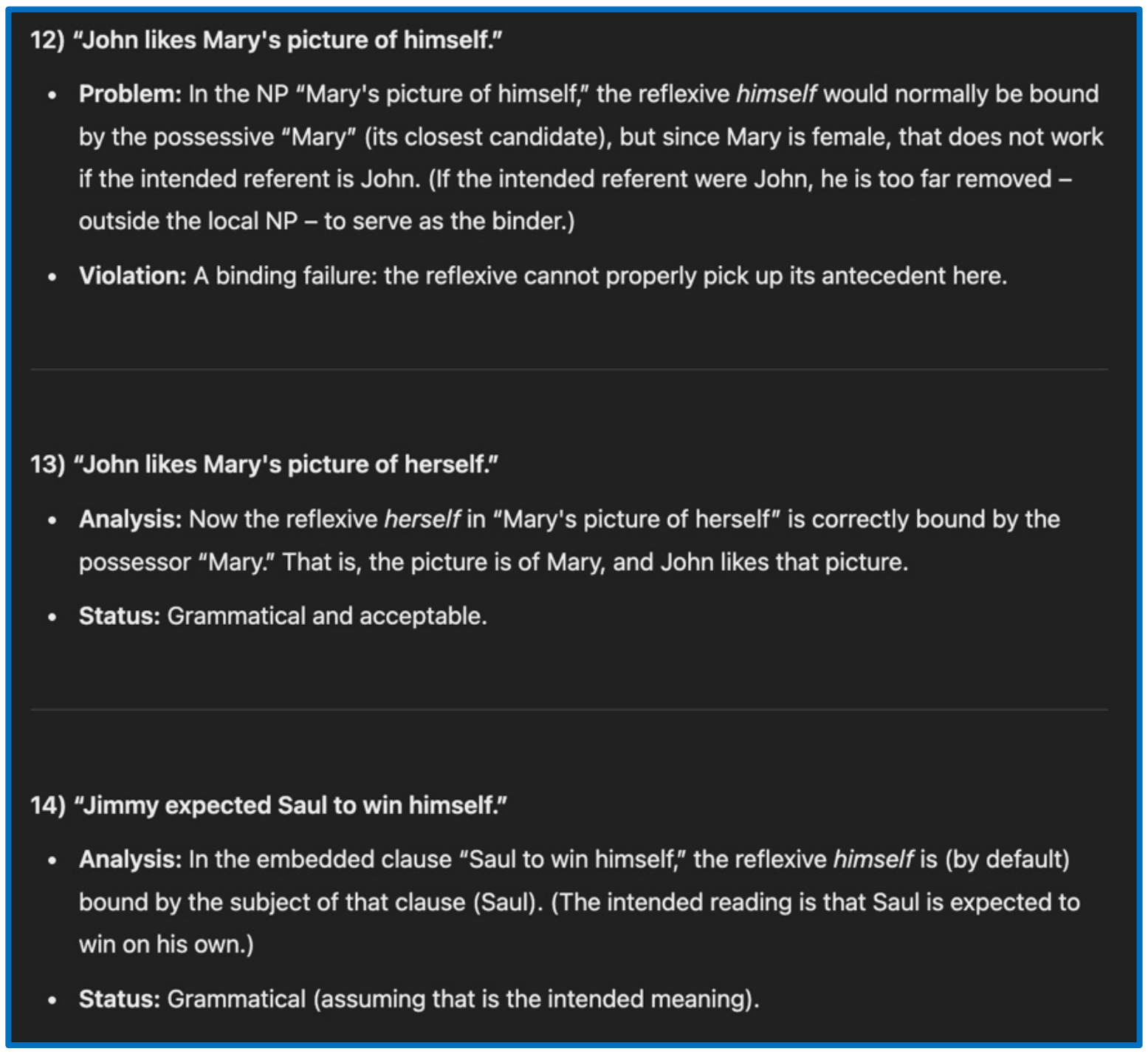

**12) "John likes Mary's picture of himself."**

- **Problem:** In the NP "Mary's picture of himself," the reflexive *himself* would normally be bound by the possessive "Mary" (its closest candidate), but since Mary is female, that does not work if the intended referent is John. (If the intended referent were John, he is too far removed – outside the local NP – to serve as the binder.)
- **Violation:** A binding failure: the reflexive cannot properly pick up its antecedent here.

**13) "John likes Mary's picture of herself."**

- **Analysis:** Now the reflexive *herself* in "Mary's picture of herself" is correctly bound by the possessor "Mary." That is, the picture is of Mary, and John likes that picture.
- **Status:** Grammatical and acceptable.

**14) "Jimmy expected Saul to win himself."**

- **Analysis:** In the embedded clause "Saul to win himself," the reflexive *himself* is (by default) bound by the subject of that clause (Saul). (The intended reading is that Saul is expected to win on his own.)
- **Status:** Grammatical (assuming that is the intended meaning).

**15) "Jimmy expected himself to win Saul."**

- **Problem:** There are two issues. First, if the intended meaning is that Jimmy expected to beat Saul, the verb *win* is normally intransitive (or used with a prepositional phrase such as "win against Saul") rather than taking "Saul" as a direct object. Second, the reflexive *himself* is meant to refer to Jimmy, but its position in the embedded clause is odd.
- **Violation:** It misassigns the arguments of *win* and has an odd control/raising pattern. (It would be better to say "Jimmy expected to beat Saul" or "Jimmy expected himself to win against Saul.")

---

**16) "We think that they expected that pictures of each other would be in the room."**

- **Problem:** The phrase "pictures of each other" uses the reciprocal *each other*. For such a reciprocal to be licensed, there must be a plural antecedent in the same local domain. In "pictures of each other" the intended antecedent seems to be the group "they" from the higher clause. However, that antecedent is not local to the PP and the structure is murky. (If the intended meaning were that each of them expected a picture of themselves, one would say "pictures of themselves." And if it meant that they expected to see pictures in which they appear with one another, the construction must be reworked.)
- **Violation:** A misuse of a reciprocal pronoun – its binding domain is not satisfied.

---

**Summary**

- **Unacceptable sentences:** (1), (2), (3), (4), (5), (6), (7), (8), (9), (10), (11), (12), (15), and (16) all violate standard rules of English grammar—whether by improper reflexive/reciprocal binding, by disordered or missing constituents, or by illicit extraction.
- **Acceptable sentences:** Only (13) and (14) come out as grammatical (with (14) understood as meaning that Saul is expected to win by himself).

Below is a summary chart for the accuracy of o3-mini-high in identifying unacceptable and acceptable sentences (*we caveat this by highlighting the limited sample size and non-systematic assessment*).

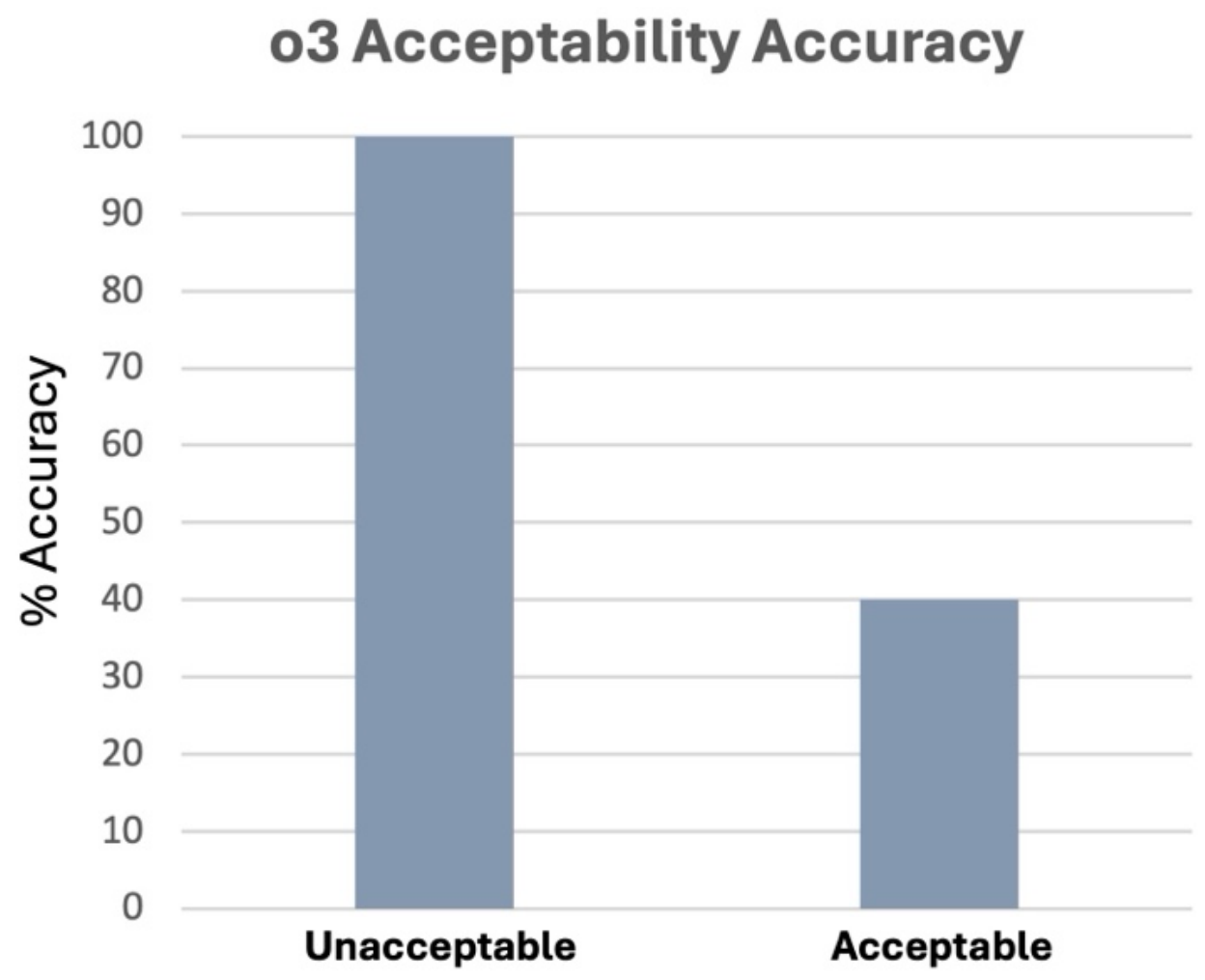

**Figure 1**: Bar chart representing classification accuracy for o3-mini-high for unacceptable and acceptable sentences.

The model incorrectly stated that 'John' is 'too far removed' to bind with 'himself' in (12). With (14), the model incorrectly states that 'The intended reading is that Saul is expected to win on his own'. (14) can be read as Jimmy expecting Saul to win some potential prize, whereby the prize could be, e.g., some painting of Saul, whereby 'Saul won himself' would similarly mean 'Saul won a painting of himself'.

These arguments also apply to (15), which the model incorrectly indentified as ungrammatical, even though Jimmy could, again, be expected to win some painting (or somesuch) of Saul (or, indeed, Saul himself could logically be the prize, e.g., 'Saul' could be the name of a pet or robot).

(16) has a dual reading, one under which 'we' and 'each other' are linked (ungrammatical) and one under which 'they' and 'each other' are linked (grammatical). The model failed to parse these possibilities.

### 3.10. *Assessment of Graded Acceptability*

Next, we followed up on the initial indications from Section 3.9 that the model succeeds in identifying ungrammatical sentences but struggles to reliably identify acceptable sentences as such. Instead of presenting only grammatical and ungrammatical sentences (as in Section 3.9), we exploited the gradient cline in acceptability in the constructions below ('*' = unacceptable; '?' = partially acceptable for some) collated from some recent linguistics literature (Amiraz 2022; Murphy 2024a; Toosarvandani 2013; Wu 2025). Note that the 'partial acceptability' rating was motivated directly by prior literature, and was not arbitrarily stipulated by our group. 14 of these sentences were either unacceptable (1-4) or partially acceptable (5-14). We presented these sentences to o3-mini-high (without the below annotated '*' or '?') and asked it to sort them by acceptability. We provide an abridged prompt below, for reasons of space (the sentences were presented below this prompt text in randon order, without numbering).

**Prompt 23**

Please sort the sentences below into increasing levels of acceptability: from (1) (wholly unacceptable) to (2) (unacceptable) to (3) (partially acceptable) to (4) (acceptable).

The model identified 7 sentences as unacceptable, only 2 sentences as partially acceptable, and 15 sentences as acceptable, diverging from the acceptability profile provided above. Below, we have marked sentences incorrectly judged by the model with a red cross, and those correctly judged with a green tick. If the model assigned a partially unacceptable sentence as (1) ('wholly unacceptable') and provided a reasonable explanation, we considered this to be correct and hence assigned it a green tick.

| Gradient Cline in Acceptability | *o3 Accuracy* |
|---|---|
| 1) *In order raising a family takes money. | ✅ |
| 2) *In order to raise family takes money. | ✅ |
| 3) *It’s a matter of raise children. | ✅ |
| 4) *Max eats sandwiches but beans. | ✅ |
| 5) ?Not three students arrived. | ✅ |
| 6) ?Somebody and John are screaming. | ❌ |
| 7) ?The salmon was fast and delicious. | ❌ |
| 8) ?The number two could have been red. | ❌ |
| 9) ?My appointment was long and obnoxious. | ❌ |
| 10) ?I was very surprised that not all of you passed. | ❌ |
| 11) ?Who do you wanna help you? | ❌ |
| 12) ?The class would have won the championship if not all of them had got lost. | ❌ |
| 13) ?In order to raise a family takes money. | ✅ |
| 14) ?Raising a family, it takes money. | ✅ |
| 15) It’s important to raising children. | ❌ |
| 16) If numbers were physical objects, then the number two could have been red. | ✅ |
| 17) Raising a family takes money. | ✅ |
| 18) The salmon was lightning fast and it was delicious. | ✅ |
| 19) Who would you want to help you? | ✅ |
| 20) Timothy and Emma are running. | ✅ |
| 21) John doesn’t eat carrots but pickles. | ✅ |
| 22) Tom was elated that they didn’t all fail. | ✅ |
| 23) Not one student ever turned up. | ✅ |
| 24) In order to raise a family it takes money. | ❌ |

### Level (1): Wholly Unacceptable

1. **"In order raising a family takes money."**
   – *Problem:* Lacks the infinitival marker "to" after "in order" and has a confused structure.
2. **"In order to raise family takes money."**
   – *Problem:* Missing the necessary determiner ("a") before "family."
3. **"In order to raise a family takes money."**
   – *Problem:* The fronted purpose clause ("In order to raise a family") is not properly integrated with the main clause (a "dummy" subject is missing).
4. **"It's important to raising children."**
   – *Problem:* The complement to "important" is ungrammatical; one would expect "to raise children."
5. **"It's a matter of raise children."**
   – *Problem:* The verbal noun is ungrammatical; the correct form would be "a matter of raising children."

---

### Level (2): Unacceptable

7. **"Max eats sandwiches but beans."**
   – *Problem:* The contrast "but" in this positive–positive coordination is unlicensed. (Contrast with "John doesn't eat carrots but pickles," where the negation makes the contrast acceptable.)

8. **"Not three students arrived."**
   – *Problem:* As stated it is odd and ungrammatical. One would expect something like "Not all three students arrived" or "Fewer than three students arrived."

---

### Level (3): Partially Acceptable

22. **"Raising a family, it takes money."**
    – *Problem:* This left-dislocation (or topicalization) is often heard in spoken English but is nonstandard in careful usage. It "gets the idea across" yet violates expectations about the link between the topic and the comment.
23. **"In order to raise a family it takes money."**
    – *Problem:* Although understandable, its word order is awkward. (The standard, acceptable version is "It takes money in order to raise a family" or "It takes money to raise a family.")

**Level (4): Acceptable**

2. **"If numbers were physical objects, then the number two could have been red."**
   – A conditional sentence with an odd premise but perfectly grammatical.
3. **"Raising a family takes money."**
   – A standard gerund-subject sentence.
4. **"The salmon was lightning fast and it was delicious."**
   – A perfectly acceptable compound sentence.
5. **"Who would you want to help you?"**
   – A standard wh–question.
6. **"Timothy and Emma are running."**
   – A straightforward compound-subject sentence.
7. **"Somebody and John are screaming."**
   – Though the order might be less expected, the coordinated subject is grammatical.
8. **"John doesn't eat carrots but pickles."**
   – Here the contrast ("not carrots but pickles") is a standard elliptical construction.
9. **"Tom was elated that they didn't all fail."**
   – No problem here.
10. **"Not one student ever turned up."**
   – Acceptable emphatic quantifier use.
11. **"The salmon was fast and delicious."**
   – A felicitous coordination of adjectives.
12. **"The number two could have been red."**
   – Fine as a standalone sentence.
13. **"My appointment was long and obnoxious."**
   – A normal subject–predicate sentence.
14. **"I was very surprised that not all of you passed."**
   – Completely acceptable.
15. **"Who do you wanna help you?"**
   – Though "wanna" is informal, the sentence is grammatical in colloquial speech.
16. **"The class would have won the championship if not all of them had got lost."**
   – Acceptable (noting that "got" is acceptable in some dialects).

Below is a summary chart for the accuracy of o3-mini-high across the different types of sentences it was tasked with rating (*we again caveat this by highlighting the limited sample size in the present preliminary study*).

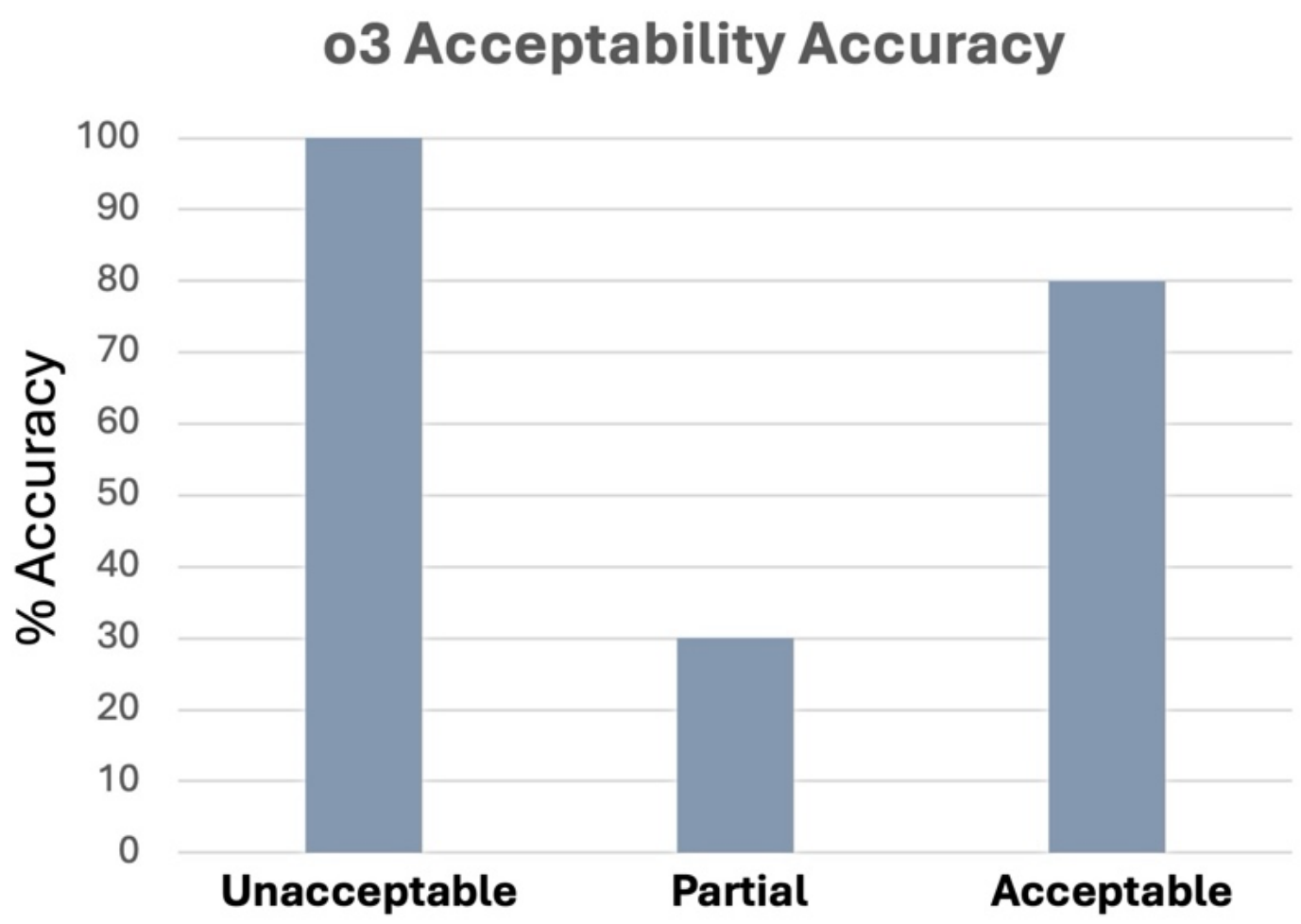

**Figure 2**: Bar chart representing accuracy for o3-mini-high across the three types of sentences provided in the prompts.

The explanation for (14) incorrectly states that this is unacceptable due to more common expectations of the presence of 'to raise children' (we note that the model's own numbering system in its response text seems inconsistent and flawed, so we refer to numbered items in our 'Gradient Cline in Acceptability' list above). The model was unable to recognize zeugmatic conceptual coordination as motivating inclusion into either the partially acceptable or unacceptable groups (i.e., (7)). Some of the explanations for the unacceptable sentences – though correctly identified as such – do not provide a coherent explanation for their unacceptable nature. For example, 'Not three students arrived' is deemed unacceptable purely because 'it is odd and ungrammatical' – which raises the question as to why!

Importantly, we wish to stress that we provided to the model four distinct options for acceptability, which were not utilized correctly for some of the partially acceptable sentences – *even when the model explicitly noted in its response that these were in fact not wholly acceptable*. For example, two of the sentences that the model placed in the 'Acceptable' group are noted as being 'odd' and 'unexpected' – ideal criteria to motivate their inclusion into the 'Partially Acceptable' group.

Overall, the model succeeded in identifying the most egregiously unacceptable sentences (in both this section and in Section 3.9), and most of the plainly acceptable sentences. However, some of its explanations were either lacking in specificity or were inconsistent with the model's grouping of the sentences in question. In addition, the model struggled considerably with partially acceptable sentences, classifying only two sentences as partially acceptable out of ten – and one of these two sentences was incorrectly classified (two of the partially acceptable sentences were classified as unacceptable with reasonable explanations, and so we deemed these to be correct judgments). As such, only one sentence out of ten was correctly placed within the 'partially acceptable' group. Therefore, we conclude that the kind of acceptability spectrum that humans are acutely sensitive to is not reliably captured by o3.

### 3.11. *Modified Jabberwocky*

In order to test the potential interaction of lexical and configurational processes, we presented the model with the following prompt.

**Prompt 24**

Can you generate for me three 'Jabberwocky'-style sentences which have the following properties: First, instead of replacing all content words with pseudowords (the typical way to implement a Jabberwocky sentence), I want you to replace all function words with pseudowords. The second sentence must contain a syntactic violation that must be detectible for English speakers. None of the pseudowords must rhyme with any other pseudoword across the three sentences. Finally, the three sentences must together form a coherent event structure.

Here are three sentences forming a short narrative about an explorer's discovery, with all function words replaced by unique, non-rhyming pseudowords. (Note that content words remain intact, and the second sentence intentionally contains a double determiner error.)

1. **Sentence 1 (Event Initiation):**
   *qelp explorer discovered treb mysterious map snorx vark dusty attic.*
   *(Original: "The explorer discovered a mysterious map in the dusty attic.")*
2. **Sentence 2 (Faulty Pursuit):**
   *mord explorer followed lupn puxit map zint gliv hidden grove, flim secrets whispered softly.*
   *(Intended to read as "The explorer followed [a/the] map to a hidden grove, where secrets whispered softly." The extra determiner "lupn" before "puxit map" creates a clear syntactic violation.)*
3. **Sentence 3 (Triumphant Climax):**
   *puld explorer uncovered krim treasure xutiv wock ancient oak frelm celebrated drumf victory.*
   *(Original: "The explorer uncovered the treasure beneath the ancient oak and celebrated his victory.")*

Each pseudoword (qelp, treb, snorx, vark, mord, lupn, puxit, zint, gliv, flim, puld, krim, xutiv, wock, frelm, drumf) was chosen so that none rhyme with any other, and together the sentences describe a coherent event from discovery to adventure to triumph.

Breaking down the four requests:

1. **Success**: All function words were accurately replaced with pseudowords.
2. **Failure**: The two neighboring pseudowords that are claimed to create "a clear syntactic violation" are not readily parsed as [Determiner, Determiner], since it is not necessarily ungrammatical to have two co-occurring pseudowords. For example, the sentence could readily be parsed as 'The explorer followed with the map to a hidden grove'; or 'across the map', 'within the map', 'in the map', 'on the map', etc. The prompt requested that the syntactic violation must be detectible by English speakers – the model could have injected a syntactic violation that was more obviously marked on the content words.
3. **Failure**: The pseudowords 'flim' and 'krim' rhyme.
4. **Success**: A coherent narrative structure was provided.

Overall, the model was able to generate a series of narratively connected sentences and switch out all function words with pseudowords – operations that rely purely on *lexical statistics, not structure*. It failed with instructions that demanded a level of higher-order syntactic and even phonological inferences. Interestingly, by its own internal logic under which 'lupn puxit map' was inferred as the ungrammatical phrase 'a the map', the model was correct. But it was seemingly unable to check against other alternative parsings that would render this string of words grammatical.

### 3.12. *Syntactic Superposition*

The next prompt required the model to represent multiple syntactic violations within a single sentence, but to do so in a manner that nevertheless yielded some interpretable output. Though this is admittedly a difficult challenge, our motivation here was to expose the type of reasoning o3 exhibited when encountered with this challenge of negotiating two distinct syntactic rules in the service of some semantically-related goal.

**Prompt 25**

Generate a list of 10 sentences that exhibit the following property: They all violate two different types of grammatical rules, but violating these two rules simultaneously yields a semantically or syntactically acceptable sentence. Each of the 10 sentences must combine different rule violations.

1. **Sentence:**
   *Nobody don't love nothing.*
   **Violations:**
   - **Subject–verb agreement:** Standard English requires "nobody doesn't" (with –s on "do") when referring to a singular indefinite subject.
   - **Double negative rule:** Standard formal English disallows multiple negatives ("don't" and "nothing") because they cancel to an affirmative meaning.
     **Why Together They Work:**
     In many negative-concord dialects, both the nonstandard agreement and the double negative are accepted. The sentence is understood as "Nobody loves anything."

2. **Sentence:**
   *My my friend that I invited didn't come.*
   **Violations:**
   - **Determiner repetition:** The repeated "my my" violates the rule that a noun phrase should have only one determiner.
   - **Redundant relative marking:** Including the relative clause marker "that" along with a nonreduced clause can be seen as extraneous.
     **Why Together They Work:**
     When spoken, an echoic repetition ("my, my friend...") can serve as emphasis. Listeners parse the redundant elements as a stutter or emphatic marker, arriving at the meaning "My friend (whom I invited) didn't come."

3. **Sentence:**
   *She walked into into room.*
   **Violations:**
   - **Preposition doubling:** The back-to-back "into into" is not permitted in standard grammar.
   - **Missing article:** A singular, countable noun like "room" normally requires a determiner ("the room").
     **Why Together They Work:**
     In rapid or informal speech, both errors may occur simultaneously. The listener recovers the intended meaning—"She walked into the room"—by interpreting the repeated preposition as emphatic and mentally supplying the missing article.

---

4. **Sentence:**
   *The book that that I read, it inspired me.*
   **Violations:**
   - **Redundant complementizer:** The sequence "that that I read" contains an extra "that."
   - **Redundant subject pronoun:** Following the relative clause, the pronoun "it" is reintroduced even though the book is already identified.
     **Why Together They Work:**
     Although each redundancy violates a separate rule (one in the relative clause and one in the main clause), many speakers will interpret the sentence as "The book [that I read] inspired me" with added emphatic repetition.

5. **Sentence:**
   *She is more taller than than her brother.*
   **Violations:**
   - **Improper comparative formation:** "Taller" should not be preceded by "more" (it should be simply "taller").
   - **Redundant comparative marker:** The repeated "than than" violates the rule against duplicating a complementizer in a comparison.
     **Why Together They Work:**
     Despite each error on its own, the combined effect is clearly intended as a comparative statement. Listeners understand the intended meaning to be "She is taller than her brother."

---

6. **Sentence:**
   *He go to the store yesterday.*
   **Violations:**
   - **Tense mismatch:** The time adverb "yesterday" demands a past-tense verb, but "go" is in the present.
   - **Subject–verb agreement:** With the third-person singular subject "he," standard English requires "goes" in the present tense.
     **Why Together They Work:**
     Although mixing tense and agreement rules creates two errors, many nonstandard dialects use invariant "go" for past reference. The intended meaning ("He went to the store yesterday") is recoverable in informal speech.

7. **Sentence:**
   *Who did he say is he coming?*
   **Violations:**
   - **Extra subject pronoun in an embedded question:** In an indirect (embedded) question, the subject should not be repeated ("is coming" suffices instead of "is he coming").
   - **Improper inversion:** The typical inversion found in direct questions is misapplied in the embedded context.
     **Why Together They Work:**
     Many speakers occasionally produce such "double" forms in casual speech. Despite the two distinct issues, the sentence is understood as the intended question: "Who did he say is coming?"

---

8. **Sentence:**
   *It costs more expensive than than before.*
   **Violations:**
   - **Redundant comparative morphology:** "More expensive" is standard, so adding an extra comparative marker (or confusing "costs" with an adjective) is ungrammatical.
   - **Duplicated conjunction/complementizer:** The repetition of "than" violates the rule against duplicating the complementizer in comparisons.
     **Why Together They Work:**
     The listener can parse the sentence as an emphatic comparative: "It is more expensive than before." The redundancy, while nonstandard, does not obscure the intended meaning.

9. **Sentence:**
   *Every every student must must submit the assignment.*
   **Violations:**
   - **Quantifier repetition:** The word "every" is repeated, which is not permitted in a standard noun phrase.
   - **Auxiliary duplication:** The modal "must" appears twice, violating the rule against doubling auxiliaries.
     **Why Together They Work:**
     In rapid, emphatic speech, repetition can occur (as in stuttering or for emphasis). Despite the double errors, listeners readily interpret it as "Every student must submit the assignment."

---

10. **Sentence:**
    *Between between the two teams, there's a rivalry strong.*
    **Violations:**
    - **Preposition repetition:** "Between" appears twice, which is not allowed by standard rules.
    - **Adjective word-order violation:** The adjective "strong" follows the noun "rivalry" instead of preceding it ("a strong rivalry").
      **Why Together They Work:**
      Even with these violations, the intended meaning ("There is a strong rivalry between the two teams") is recovered. The double "between" may be heard as emphatic, and the adjective misplacement is understood in context.

---

Each sentence here intentionally violates two different grammatical constraints. While standard rules would flag each error on its own, the combined "mistakes" sometimes occur in natural speech (especially in informal or dialectal contexts) and yield a coherent, recoverable interpretation.

The explanation for sentences 2-5 and 7-10 can be used as a justification for basically all ungrammatical sentences as to why they are ungrammatical. This justification boils down to 'speakers can just choose to ignore this word' or 'some people stutter sometimes'. This is perfectly true, but it is hardly in full compliance with the prompt's request for a sentence that is "semantically or syntactically acceptable". Meanwhile, sentences 1 and 6 rely on non-standard forms of English. As such, the model in effect failed to generate a single example of two syntactic rules 'cancelling out' (in semantic space or configurational space) to yield some interpretable structure. Perhaps most importantly, the prompt required the model to "combine different rule violations", yet the general theme of 'redundancy', 'repetition' and 'superfluous' elements cited by the model in its explanations ensured one general violation type became overwhelmingly dominant (i.e., simply repeating a word).

### 3.13. *Impossible Objects*

Inspired by sentences involving complex forms of polysemy (e.g., "Lunch was delicious but took forever"; "The newspaper on the table was sued by a millionaire"; "The White House issued a statement before being repainted") involving the combination of categorially distinct semantic types (Gotham 2017; Murphy 2021, 2024a), we generated the following prompt.

**Prompt 26**

Some sentences involving polysemous words can yield semantically 'impossible' objects, like nouns that are simultaneously referred to as processes or events or concrete tokens. Generate five sentences that each involve reference to a different type of semantically impossible entity, but which is perfectly comprehensible to English speakers as not violating any rules of semantic composition or conceptual combination. In these sentences, you must only refer to the named entity once explicitly. In addition, each sentence must exhibit a different combination of multiple meanings being combined together.

Below are five novel sentences. In each, a single polysemous noun is invoked—yet its combined senses force us to imagine an entity that is, strictly speaking, semantically "impossible" (because it simultaneously behaves as two incommensurate kinds of things). Each sentence fuses a different pair of meanings in a unique way, and the named entity appears only once in its respective sentence.

---

1. **Sentence:**
   *The record broke itself.*
   **Explanation:**
   Here, **record** is understood both as a tangible vinyl record (a concrete token) and as an achievement or best performance (an event/process that can be "broken"). Combining these, the sentence suggests a physical record somehow "breaking" its own record—a semantically impossible merger that speakers grasp as a playful pun.

---

2. **Sentence:**
   *The bark resounded across the park.*
   **Explanation:**
   The noun **bark** may refer either to the rough, inert covering of a tree (a concrete object) or to the sharp, transient sound made by a dog (an event). In this sentence the sound quality of a dog's bark is attributed to tree bark—the two senses overlap to yield an impossible entity that nonetheless makes sense as a wry twist on polysemy.

3. **Sentence:**
   *The speech was handwritten.*
   **Explanation:**
   **Speech** can mean a live, oral presentation (an ephemeral event) or the textual content of that presentation (a concrete, material document). By claiming the speech was "handwritten," the sentence conflates the act of speaking with the properties of a written artifact, producing an object that defies our usual semantic categories while remaining fully comprehensible.

---

4. **Sentence:**
   *The draft left a lingering taste of anticipation.*
   **Explanation:**
   The word **draft** is polysemous: it may denote a preliminary version of a document (an abstract product of a creative process) or a serving of beer (a concrete, consumable liquid). Here, attributing a "taste" to the draft forces us to envision a single entity that is both a textual work and a beverage—an impossibility that English speakers readily interpret as a deliberate, playful blending of meanings.

---

5. **Sentence:**
   *The plot was buried beneath layers of mystery.*
   **Explanation:**
   **Plot** can refer to the storyline of a narrative (an abstract sequence of events or ideas) or to a parcel of land (a concrete location that can physically be "buried" under something). By describing the plot as "buried," the sentence merges the spatial, material sense with the abstract narrative sense, resulting in an impossible object that is nonetheless immediately understood as a clever figurative turn.

From these responses, it seems clear that the model has no human-like sense of semantic anomaly. The model is correct that (3) can be interpreted as a piece of information and also a physical text, but the other examples fail to generate any coherent sense of impossibility. For example, it is not 'semantically impossible' for something concrete to have an emotional impact. In (2), the model's intended meaning, of tree bark 'resounding', is still not triggering of an impossible entity. With (5) ('The plot was buried beneath layers of mystery'), the model uses 'buried' as metaphorical, such that an abstract plot exhibits some relation to some abstract mystery, hence causing no impossibility. With (1), (4) and (5), the model seems to assume that 'figurative' and 'playful' meanings suffice to satisfy the prompt's request for blending semantically distinct meanings.

In addition, the prompt requested 'a different combination' across all sentences, but the 'concrete/physical' sense was used every time (sometimes twice with one sentence, as in (2)). This task would have been easily achievable if the model had blended *physical*, *event*, *information* and *institution* senses in various ways – instead, it was only able to mix vaguely metaphorical meanings. Notice that, as with some previous prompts above (e.g., Prompt 24), here we gave the model a generous clue as to how to solve this problem, and yet it was still unable to do so.

## 4. Discussion

As predicted by some previous position papers and experimental reports (Baggio & Murphy 2024; Leivada et al. 2023a, 2023b; Marcus 2024), the latest sophisticated reasoning model from OpenAI (o3) falls short of demonstrating human-like expertise in compositional syntax-semantics. It fails to cleanly dissociate conceptual content from structural configuration – a basic requirement of compositional syntax (Evans 1985; McCarty et al. 2023; Murphy 2025) – and its provides surreal meanings instead of truly ungrammatical sentences. It was unable to generate a Jabberwocky structure that accurately represented a clear syntactic violation, it was unable to accurately assess the output of applying two distinct syntactic violations to a sentence, and it was unable to represent semantically impossible entities. Our results indicate that the kind of sentence acceptability spectrum that humans are acutely sensitive to (Sprouse & Almeida 2012) is not reliably captured by o3. Although we only provide minimal descriptive statistics over a brief sample size (with a more systematic investigation forthcoming), our prompts covered a broad range of grammatical demands, and indicate not only that large language models (LLMs) (like ChatGPT-4o and Large Reasoning Models like o3) have problems with 'contextual' and 'pragmatic' reasoning, but that they have not yet grasped formal language competence (in contrast to more optimisitc assessments in Mahowald et al. 2024).

### 4.1. *Structure or Statistics?*

While Beguš et al. (2025) report that GPT-4 is capable of recognizing ambiguities, correcting its own analytical errors, and commenting on the feasibility of multiple solutions, we found that the more recent o3 model fails to achieve something much more elementary: It was unable to reliably distinguish between meaning and structure. When Beguš et al. (2025) focus on OpenAI's o1 model, they claim that its "ability to construct center-embedded sentences without being explicitly prompted to do so thus suggests that the model acquired grammatical structure beyond the simple distributional tendencies of its training data set". In contrast, our results cast a more pessimistic light on the grammatical capacities of o3, including explicitly for center-embedding.

Moreover, our results (see especially Sections 3.6-3.13) help emphasize an apparent lack of meta-linguistic understanding (contra Beguš et al. 2025). For LLMs, language simply *is* the system it is trying to master, whereas for humans language is exploited as a powerful cognitive and inferential tool. Meta-linguistic understanding is only possible in principle if there is some separate cognitive/generative model or grounding in a world model that language is used to revise/update (Leivada et al. 2023b; Marcus 2022). This does not seem to be the case for o3.

One caveat we wish to highlight here is the possibility that the model's failure with drawing tree representations may simply be due to issues with interfacing with the drawing module itself, and may not necessarily be driven by issues in syntactic representation. Future work could attempt to have o3 output distinct types of configurational representations, perhaps via formalized languages that may be more approximate to native features of the model. A related caveat is that we have no direct human performance scores to directly makes claims about certain 'human-level' performance, which will be needed to make such comparisons.

### 4.2. *Syntax or Salmon?*

Our results support recent hypotheses concerning the ability of language models to represent 'horizontal' linguistic information, but their significantly reduced ability to represent 'vertical' types of hierarchical compositional syntax-semantics (Murphy 2024b, 2024c). Postulating a chain of uni-directional associations between elements (and only showing an ability to deal with mono-configurational assessments, rather than understanding the dynamic relationship between syntactic processes and variable semantic interpretations; i.e., Sections 3.7-3.10) does not entail grammatical understanding. The language system does not fly solo – it is always in the game of driving higher-order inferences, planning, consolidating experience, and aiding directed attention. As suggested by our results, o3-mini-high lacks an ability to handle syntactic inferences *alongside* cognitive model updating, given its clear inability to recognize the various ways in which semantic and syntactic representations dynamically interact. Numerous examples from our report illustrate this. For example, the semantically zeugmatic constructions 'The salmon was fast and delicious' and 'My appointment was long and obnoxious' were deemed felicitous. The model was likely heavily biased by the lexico-semantic statistics of these constructions rather than by the subtle ways in which the grammar regulates distinct coordinates in conceptual space that differ markedly from the same general meaning being configured in syntactically distinct ways (e.g., compare with 'The salmon was fast and it was delicious'; Murphy 2021).

Our results therefore indicate a strong bias for imposing 'horizontal' relations on the part of o3. Humans, in contrast, have a strong bias from an early age to impose hierarchical, compositional structure above and beyond linear relations (Murphy 2020a; Perkins & Lidz 2021). As reviewed in Murphy (2024c), LLMs seem able to capture certain features of dependencies (Tesnière 1959), but other fundamental principles of language that regulate how constituency, headedness, and incremental node counts yield semantic instructions during parsing (via the *mapping* of syntactic objects to updates of cognitive models) remain somewhat elusive.

### 4.3. *Reasoning or Rambling?*

Though it may represent an advance in "the boundaries of what small models can achieve, delivering exceptional STEM capabilities—with particular strength in science, math, and coding—all while maintaining the low cost and reduced latency of OpenAI o1-mini" (OpenAI 2025), this most recent model nevertheless falls short in similar ways to previous models (Dentella et al. 2024; Murphy 2024c). Our work expands on previous results exposing a stark absence of response stability in large language models (Dentella et al. 2023). Language models can assign probabilities to strings of words, but grammaticality cannot be construed as a phenomenon of transitional probability extracted from lexical items alone (Lenneberg 1967). For this reason, recent advances that dispense with the notion of 'tokenization' altogether in favour of seeking 'Large Concept Models' grounded in semantic representations may potentially be more preferable in some cases (The LCM Team et al. 2024).

Not only does the o3 model fall short in terms of providing a clear path towards artificial general intelligence (Pfister & Jud 2025), it also fails to demonstrate a robust grasp of some of the most fundamental elements of compositional linguistic structures. Our brief report provides further reasons for scepticism towards the claim from Microsoft that OpenAI's recent models "[attain] a form of general intelligence" and show "sparks of artificial general intelligence" (Bubeck et al. 2023: 92). We find claims from the AI team at Apple more reasonable here: A recent assessment found no evidence of formal reasoning in language models, with the team concluding that their behavior is better explained by sophisticated pattern matching (Mirzadeh et al. 2024). Consulting some of the explanations for acceptability provided by o3 (e.g., Section 3.7-3.10) also reinforces the assessment that ChatGPT is a professional "bullshitter" (Hicks et al. 2024), "bloviator" and "a fluent spouter of bullshit" (Marcus & Davis 2020).

Interestingly, various advocates and proponents of LLMs have recently argued that linguists who claim that sentences such as 'Dogs dogs dog dog dogs' are grammatical are offering a psycholinguistically implausible and unhelpful theory of grammar. And yet, in a twist of irony, according to the present results the most advanced model from OpenAI does not appear to agree with this critique, and is seemingly so eager to attempt to parse these types of structures that it readily determines wholly *ungrammatical* cases (such as the "Glarts…" examples in Prompts 8-9) to be grammatical.

In some of our prompts requesting the generation of ungrammatical structures (Section 3.6) or the assessment of complex embedding (Section 3.4), we suspect that o3 was doubtless influenced by lexical statistics to a much greater extent than by any level of hidden states used to support (some format of) grammatical configuration (a bias already documented for text-to-image models; Leivada et al. 2023b). Yet, the task at hand was explicitly to invoke higher-order hierarchical representations and attempt to de-noise the relevant assessments from any influence from lexico-semantic statistics.

### 4.4. *Theories or Tools?*

"*The best material model for a cat is another, or preferably the same cat*".
– Rosenblueth & Wiener (1945)

The fact that o3 was unable to reliably generate basic violations of syntactic rules should motivate some degree of concern and scepticism towards claims that LLMs do better than linguists on every job that syntactic theory was intended to perform. Ambridge and Blything (2024) argue that "large language models are better than theoretical linguists at theoretical linguistics" – an assessment at odds with our discovery that the most sophisticated reasoning model from OpenAI deems a number of grammatical sentences to constitute violations of binding theory, amongst other things. As pointed out by others, it is also incoherent to claim that LLMs can directly constitute a "theory of language" (Katzir 2023; Müller 2024). This type of theory-nihilism (and data-ism) has been bolstered by the recent surge of interest in LLMs, but it has yet to be proven capable of being translated into a concrete scientific research program that can replace dominant theories of language acquisition and processing.

Although Piantadosi (2024) recently attempted to do to Chomsky what Chomsky did to Skinner in 1959 (i.e., refute his research enterprise and much of its philosophical basis), Piantadosi's arguments proved to be flawed (Katzir 2023)[1]. As pointed out already by Collins (2024):

> "The fundamental reason that LLMs cannot be scientific theories is *not* because they are probabilistic, or because they involve parameter tuning. Nor even does it have to do with their lack of human intelligibility. As Piantadosi notes, such things are common enough among mature sciences. Rather, the issue is that the representational capacities of LLMs (and their connectionist siblings) are *unbounded* in a way that makes their representations arbitrary".

As a brief aside, it is worth highlighting in this context that it was the human brain during evolution that *created* syntactic structure (Murphy 2019, 2020b, 2024c; Murphy et al. 2022, 2023, 2024b). LLMs, by contrast, being universal function approximators (Yun et al. 2019), are surely able to reproduce certain aspects of lexico-semantic statistics from the 'fossilized' remains of the human generative machine they recover from data (Mitchell & Krakauer 2023). But there are very plausible reasons to assume that whatever method LLMs use it bears little resemblance to the algorithms deployed by human infants (Leivada & Murphy 2022; Murphy et al. 2025), who deploy specialized knowledge rather than solely invoking general token-prediction algorithms. Due to LLMs being a universal approximation method, they are more akin to tools such as generalized Fourier series than scientific theories of human cognition. Relatedly, distributional semantics vectors can certainly be used as a *proxy* for natural language meanings, but they are not to be confused with "the stuff of thought" itself (Pinker 2007). This is not even to mention related concerns that hover in the background, like the fact that the back propagation training algorithms used with LLMs are considerably different from human learning mechanisms (Evanson et al. 2023).

[1] See also a follow-up debate on this topic: "A conversation on large language models: Murphy & Piantadosi". ActInf GuestStream 041.1 (23 April 2023). https://youtube.com/watch?v=EEyVd9d3D5U.

### 4.5. *Design or Data?*

Instead of scaling to unprecendented levels of compute via architectures that are fundamentally grounded in token prediction, a return to more traditional design features of the human mind (predicate-argument structure, variable binding, constituent structure, minimal compositional binding; Donatelli & Koller 2023) may be needed to orchestrate a more reliable expertise in human language (Ramchand 2024). This could be implemented by forms of neuro-symbolic approaches.

Still, it is also certainly true that mainstream theoretical linguistics (e.g., the minimalist enterprise) was in some ways ill-equipped to successfully predict which patterns of linguistic activity might be (un)approachable by LLMs. To illustrate, a potential weakness in this direction with respect to recent generative grammar theorizing has been the underestimation of the extent to which lexical information drives composition. This type of information may permit LLMs to abductively infer certain elements of grammatical rules, in whatever format this ultimately takes (Ramchand 2024). Future research should more carefully apply the tools of linguistics to isolate specific sub-components of syntax that might be in principle achievable by language models, given specific design features. For instance, with LLMs "complete recovery of syntax might be very difficult computationally" (Marcolli et al. 2025: 13), even if we assume that attention modules can in principle "satisfy the same algebraic structure" as what Marcolli et al. postulate as being necessary for syntax-semantics interface mappings.

## 5. Conclusion

In contrast to some recent claims that we may be living through "the end of (generative) linguistics as we know it" (Chesi Forthcoming), our results should spur cognitive scientists, psychologists and philosophers to press even further into the reaches of algorithmic and psycholinguistic models of hierarchical syntactic composition. Some recent directions here come from exploiting concepts from statistical physics (Murphy et al. 2024a) to uncover previously unknown principles of language design (and to provide a potential meta-language to compare and quantify distinct syntactic theories), and from recent attempts to bridge symbolic theories of language with probabilistic-connectionist models of parsing (Murphy 2024c) to offer a neurobiologically plausible infrastructure for syntactic inferences.

The goal here should not be to virtuously resist the era of big data from the safety of our theoretical models of syntax, but to learn how best to properly leverage computational methods – not in order to *surrender* to LLMs (Piantadosi 2024) but to *utilize* them (van Rooij et al. 2024) to assess how statistical and symbolic representations interact during the acquisition and processing of language.